\pdfoutput=1
\documentclass[10pt, logo, twocolumn, copyright, nonumbering]{nvidiatechreport}
\usepackage{caption}
\usepackage{mdframed}
\usepackage[utf8]{inputenc} 
\usepackage[T1]{fontenc}    
\usepackage{hyperref}       
\usepackage{url}            

\usepackage{booktabs}       
\usepackage{amsfonts}       
\usepackage{nicefrac}       
\usepackage{microtype}      
\usepackage[dvipsnames]{xcolor}         
\usepackage{multirow}
\usepackage{multicol}
\usepackage{graphicx}
\usepackage{subcaption}
\usepackage[numbers]{natbib}
\usepackage{tabto}
\usepackage{xspace}
\usepackage{amsmath}
\usepackage{adjustbox}
\usepackage{enumitem}
\usepackage{wrapfig}
\usepackage{dblfloatfix}
\usepackage{algorithm}
\usepackage{algpseudocode}
\usepackage{comment}
\usepackage{cuted}
\usepackage[most]{tcolorbox}
\usepackage{xcolor}
\usepackage{graphicx}
\usepackage{hyperref}

\newcolumntype{g}{>{\color{gray}}c} 

\newcommand{\NOne}[0]{Star Elastic\xspace}

\title{Star Elastic: Many-in-One Reasoning {LLMs} with Efficient Budget Control}
\author{
Ali Taghibakhshi*,
Ruisi Cai*†,
Saurav Muralidharan*,
Sharath Turuvekere Sreenivas*,
Aditya Vavre*,
Ameya Sunil Mahabaleshwarkar,
Bilal Kartal,
Sheldon Liang,
Marcin Chochowski,
Zijia Chen,
Akhiad Bercovich,
Ran Zilberstein,
Ran El-Yaniv,
Yonatan Geifman,
Daniel Korzekwa,
Yoshi Suhara,
Oluwatobi Olabiyi,
Ashwath Aithal,
Nima Tajbakhsh,
Pavlo Molchanov
}

\correspondingauthor{ataghibakhshi@nvidia.com}

\begin{abstract}
\textbf{Abstract:}
Training a family of large language models (LLMs), either from scratch or via iterative compression, is prohibitively expensive and inefficient, requiring separate training runs for each model in the family. In this paper, we introduce \textbf{\NOne{}}, a novel LLM post-training method that adds $N$ nested submodels to a given parent reasoning model using the compute of one run ($N\times$ savings) via a single post-training job. Beyond reducing training costs, \NOne{} also addresses a fundamental limitation of efficient reasoning: the rigidity of static architectures, which forces the allocation of constant resources regardless of token difficulty. By unlocking elastic budget control, \NOne{} enables a novel inference scheme that uses different submodels for each reasoning phase (thinking and answering). \NOne{} supports (1) nesting along the SSM, embedding channel, MoE, and FFN axes, (2) learning nested submodels via an end-to-end trainable router, and (3) curriculum-based knowledge distillation. Building on the Nemotron Elastic framework~\citep{taghibakhshi2025nemotron}, we apply \NOne{} to the NVIDIA Nemotron Nano models, with a particular focus on hybrid Mixture-of-Experts (MoE) architectures: from Nemotron Nano v3 (30B/3.6A)~\citep{blakeman2025nemotron3}, we generate 23B (2.8A) and 12B (2.0A) variants with 160B training tokens. All nested models match or outperform independently trained baselines of comparable size and achieve a $360\times$ reduction versus pretraining from scratch and a $7\times$ reduction over state-of-the-art compression. Crucially, elastic budget control advances the accuracy--latency Pareto frontier, achieving up to 16\% higher accuracy and $1.9\times$ lower latency via dynamic per-phase model selection. We further extend \NOne{} to quantized regimes via Quantization-Aware Distillation (QAD), producing nested NVFP4 and FP8 elastic checkpoints that preserve zero-shot slicing while delivering substantially smaller deployment footprints.
\end{abstract}

\begin{document}
\maketitle
\begin{strip}
\vspace{-30pt} 
\begin{center}
\begin{tcolorbox}[
  colback=gray!3, colframe=gray!30, arc=2pt,
  boxsep=1pt, left=3pt, right=3pt, top=3pt, bottom=3pt,
  width=0.7\textwidth
]
\centering
\textbf{\footnotesize Elastic Models on Hugging Face} \\[2pt]

\raisebox{-0.25\height}{\includegraphics[height=1.2em]{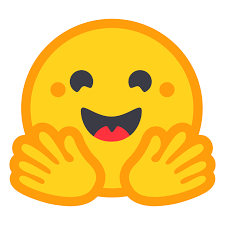}}~%
\href{https://huggingface.co/nvidia/NVIDIA-Nemotron-Labs-3-Elastic-30B-A3B-BF16}{%
  \scalebox{1.0}{\texttt{BF16}}
} \quad
\raisebox{-0.25\height}{\includegraphics[height=1.2em]{fig/hf_logo.png}}~%
\href{https://huggingface.co/nvidia/NVIDIA-Nemotron-Labs-3-Elastic-30B-A3B-FP8}{%
  \scalebox{1.0}{\texttt{FP8}}
} \quad
\raisebox{-0.25\height}{\includegraphics[height=1.2em]{fig/hf_logo.png}}~%
\href{https://huggingface.co/nvidia/NVIDIA-Nemotron-Labs-3-Elastic-30B-A3B-NVFP4}{%
  \scalebox{1.0}{\texttt{NVFP4}}
}
\end{tcolorbox}
\end{center}
\end{strip}

\section{Introduction}
\label{sec:intro}

\begin{figure*}[t]
\centering
\includegraphics[width=0.46\textwidth]{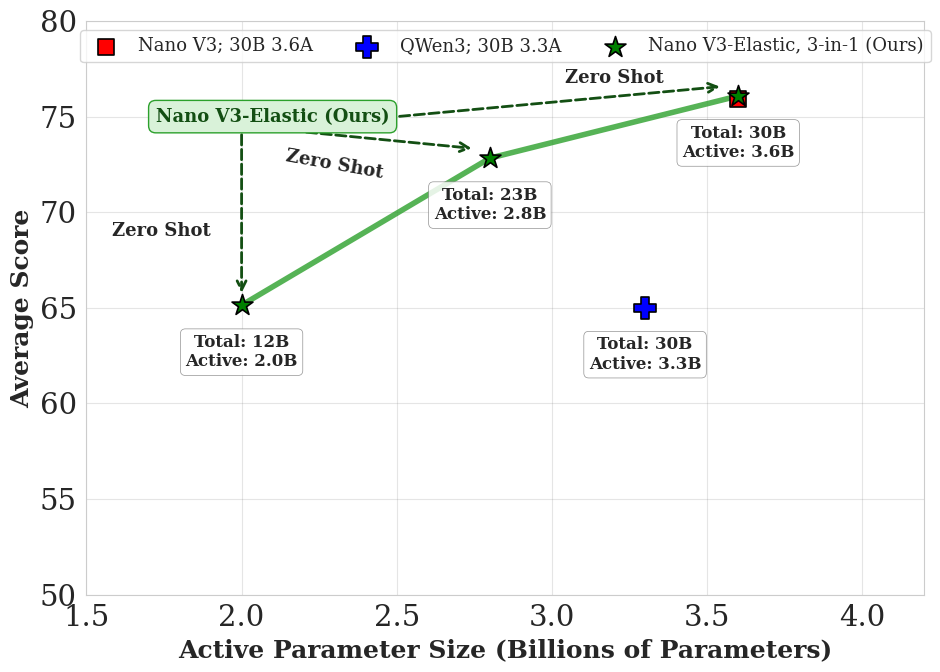}
\hfill
\includegraphics[width=0.525\textwidth]{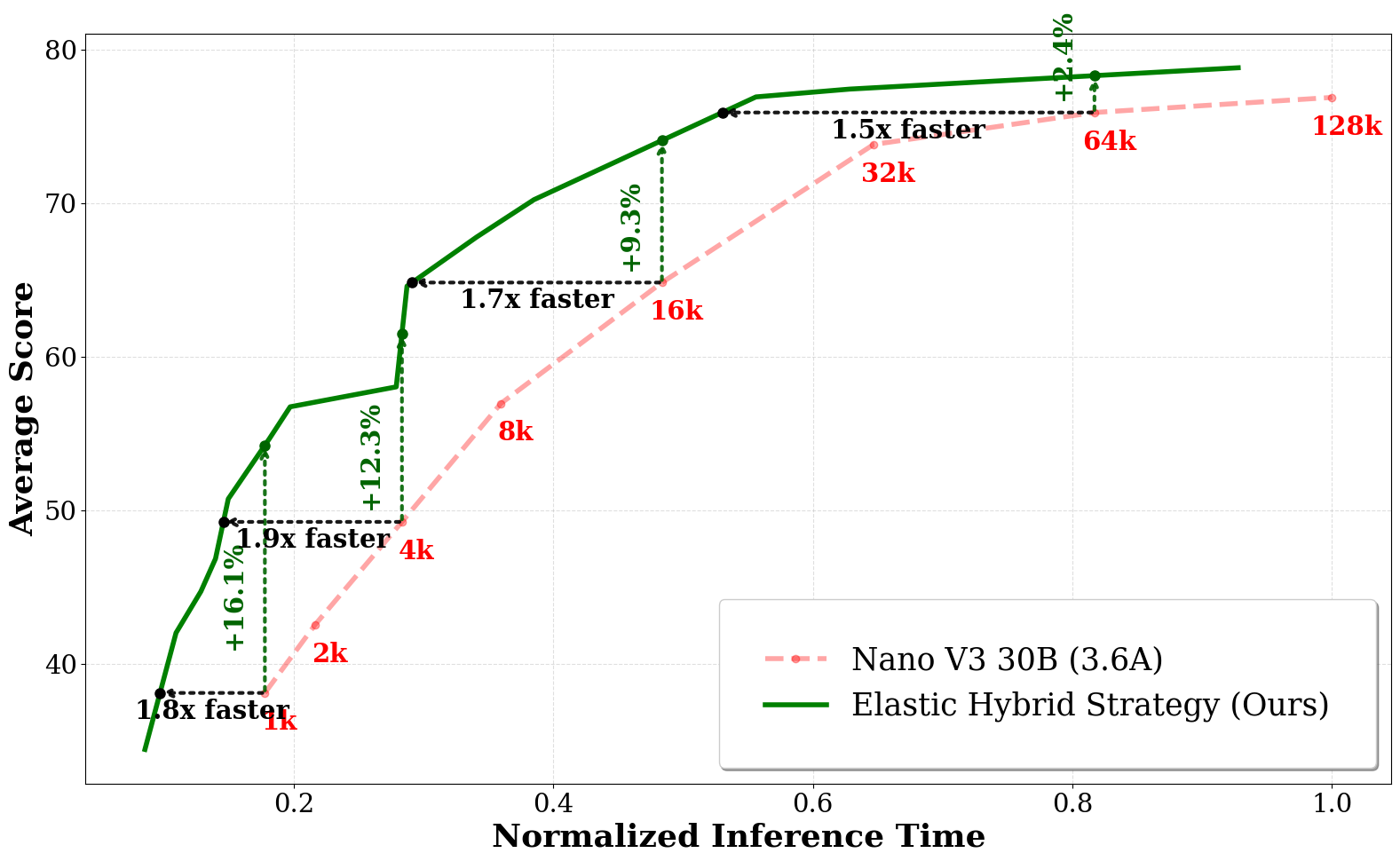}
\caption{
\textbf{Left}: Average accuracy of \NOne{} (Nano v3 Elastic) compared to parent and Qwen models across key reasoning benchmarks (AIME-2025, IFBench, GPQA, LiveCodeBench v5, and MMLU-Pro).
\textbf{Right}: Hybrid elastic budget control versus standard Nemotron Nano v3 reasoning budget control; \NOne{} improves the accuracy--speed Pareto frontier. Time is measured using vLLM.}
\label{fig:side_by_side_elastic}
\end{figure*}

Large Language Model (LLM) families typically consist of a set of fixed-size models, offering users discrete accuracy--cost trade-offs across diverse deployment settings~\citep{touvron2023llama,dubey2024llama3herdmodels}. However, each model in the family is typically trained independently and stored as a separate set of parameters, making this approach prohibitively expensive in terms of compute/storage resources, and deployment overhead. For example, the Llama-3.1 family spans 8B, 70B, and 405B parameters~\cite{dubey2024llama3}, with each variant trained from scratch on tens of trillions of tokens, multiplying training and storage costs, and restricting users to a small set of predefined model sizes.
Recent model compression methods mitigate the cost of training LLM families through structured pruning and knowledge distillation, training only the largest model from scratch and deriving smaller variants through compression~\citep{muralidharan2024compact,xia2023sheared}. While effective, these approaches still require hundreds of billions of training tokens per compressed model, keeping overall training costs high. {\em Elastic} nested models, on the other hand, embed multiple nested sub-models within a single parent model, enabling zero-shot extraction of multiple model sizes from one training run and facilitating efficient multi-size deployment~\cite{cai2024flextron,kudugunta2023matformer}. Recently, Nemotron Elastic~\citep{taghibakhshi2025nemotron} demonstrated that this paradigm can be effectively applied to hybrid Mamba--Attention reasoning LLMs, deriving 9B and 6B variants from the Nemotron Nano v2 12B parent with only 110B training tokens.

In parallel, hybrid LLMs combining attention, State Space Models (SSMs, such as Mamba), MLPs, and Mixture-of-Experts (MoE) have become popular~\citep{gu2023mamba,dao2024transformers,lieber2024jamba,glorioso2024zamba,blakeman2025nemotron}, offering improved efficiency through reduced KV cache, linear-time sequence processing, and conditional expert computation, while maintaining strong accuracy. Unfortunately, no current framework supports elasticity (nesting) across the \emph{full} hybrid Mamba--Attention--MoE design space; the limited efforts targeting compression of such models~\citep{shukla2024matmambamatryoshkastatespace} that we know of do not support heterogeneous expert or FFN channel selection, and Nemotron Elastic~\citep{taghibakhshi2025nemotron} considers only Mamba--Attention parents without MoE.

In this paper, we introduce \textbf{\NOne{}}, a post-training approach for hybrid Mamba--Transformer--MoE LLMs that produces multiple nested sub-networks at different parameter budgets from a single training run, and supports elastic budget control during inference. Our approach combines (1) importance-based ranking of embedding channels, attention and SSM heads, MoE experts, and FFN channels, (2) a learnable router that automatically determines nested submodels and employs knowledge distillation for joint sub-network optimization, (3) a two-stage training curriculum for optimal reasoning performance, (4) elastic budget control allowing dynamic per-phase model allocation, and (5) Quantization-Aware Distillation (QAD) for elastic checkpoints, yielding NVFP4 and FP8 elastic models that preserve zero-shot slicing.

We apply \NOne{} primarily to Nemotron Nano v3 MoE (30B/3.6A)~\citep{blakeman2025nemotron3}, producing 23B (2.8A) and 12B (2.0A) variants with $\sim$160B training tokens. All nested models match or outperform independently trained baselines, while enabling a $360\times$ token reduction compared to training from scratch and a $7\times$ reduction over state-of-the-art compression~\citep{taghibakhshi2025nemotron}. Crucially, elastic budget control enabled by \NOne{} advances the Pareto frontier, achieving up to 16\% higher accuracy and $1.9\times$ lower latency via dynamic per-phase model selection as shown in Figure~\ref{fig:side_by_side_elastic}. For Nemotron Nano v2 results (a 9B and 6B nested family from the 12B parent obtained with $\sim$110B tokens), as well as the corresponding training-token and deployment-memory scaling analysis, we refer the reader to the original Nemotron Elastic paper~\citep{taghibakhshi2025nemotron}.

\paragraph{Key contributions:}
\begin{itemize}
\item \NOne{} introduces the first elastic post-training method for \textbf{reasoning LLMs} with hybrid \textbf{Mamba--Attention--MoE} architectures, extending Nemotron Elastic~\citep{taghibakhshi2025nemotron} from Mamba--Attention to MoE-based hybrid models.
\item Demonstrates \textbf{elastic budget control}, dynamically allocating submodels across reasoning and answer generation phases, achieving up to 16\% higher accuracy and $1.9\times$ lower latency.
\item Introduces a \textbf{learnable router} that automatically determines nested submodel architectures, optimized through \textbf{knowledge distillation} from the parent model, and supports \textbf{heterogeneous} per-layer FFN/expert selection.
\item Achieves \textbf{significant training cost reductions} for model families: up to $7\times$ over prior compression methods and $360\times$ over pretraining from scratch.
\item Extends elastification to \textbf{quantized regimes} via Quantization-Aware Distillation, producing nested NVFP4 and FP8 elastic checkpoints that preserve zero-shot slicing while substantially reducing deployment memory.
\end{itemize}

\section{Elastic Model Construction}
\label{sec:methodology}

\begin{figure*}[t!]
\centering
\includegraphics[width=0.85\textwidth]{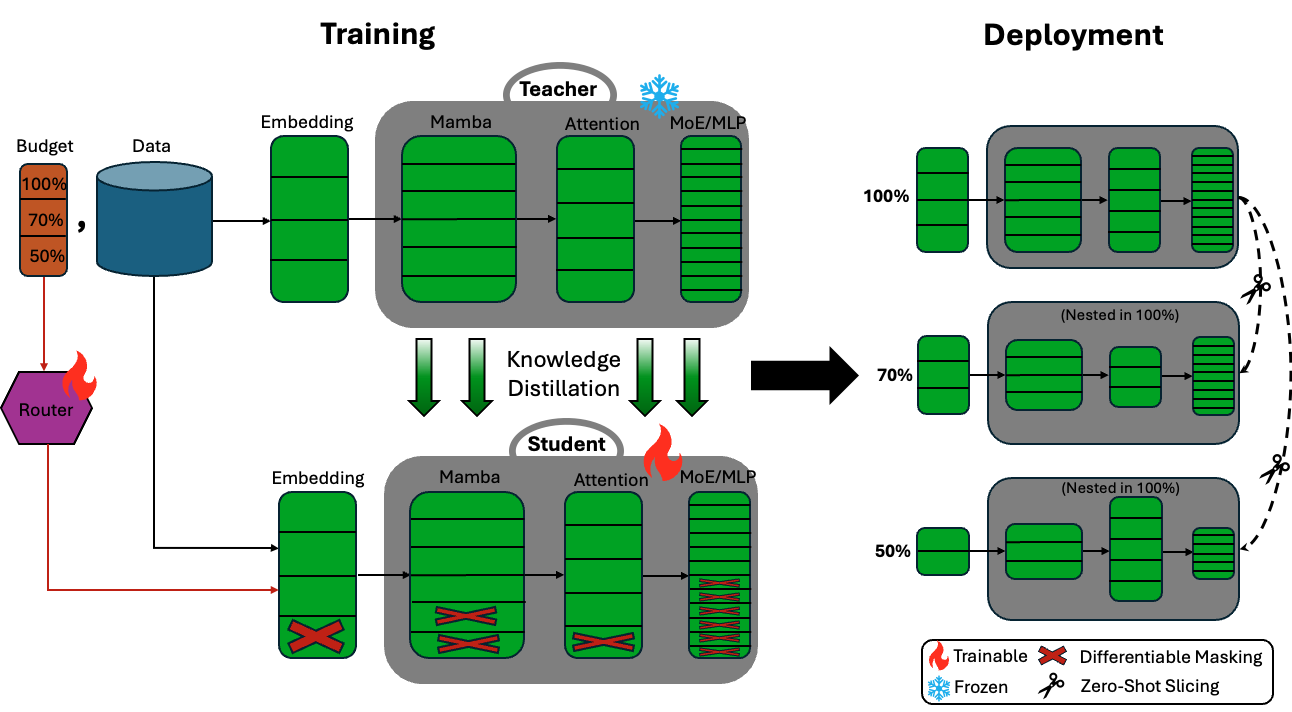}
\caption{
\textbf{Overview of the \NOne{} training and deployment pipeline.} \textbf{Training:} For each training sample, data flows to both teacher and student models. A budget (e.g., parameter count, memory, or latency target) is selected and passed to the router, which generates differentiable masks for the student model. Knowledge distillation from the model prior to elastification enables simultaneous optimization across all budget variants. \textbf{Deployment:} After training, all models are extracted zero-shot from a single elastic checkpoint: the full model and nested sub-networks are immediately available without additional training (Appendix~\ref{app:deployment-details} contains deployment details).}
\label{fig:overview}
\end{figure*}

In this section, we describe how \NOne{} converts an existing LLM into an elastic model. The pipeline consists of three stages: (1) importance estimation to rank model components (embedding dimensions, attention and SSM heads, MoE experts, and FFN channels), (2) using an elastic formulation to enable nesting along both width and depth axes, and (3) elastic training using a two-stage curriculum with a learnable router. Figure~\ref{fig:overview} showcases the \NOne{} pipeline. Several core ingredients of this pipeline---in particular, importance estimation, group-aware Mamba ranking, and the two-stage curriculum---directly build on Nemotron Elastic~\citep{taghibakhshi2025nemotron} and we refer the reader to that work for additional details, including ablations on Mamba--Attention--only parents.

\subsection{Importance Estimation and Model Preparation}
\label{sec:ranking}

The first stage of elastification involves computing the importance score for each model component, such as attention heads, FFN channels, layers, etc. Components are then sorted in decreasing order of importance, yielding ranking permutations $\sigma^{(w)}$ that reflect their contribution to model accuracy. These rankings guide the router's architecture selection during elastic training.

\paragraph{Width.} We extend activation-based importance scoring~\cite{muralidharan2024compact,taghibakhshi2025minitron,taghibakhshi2025nemotron} to hybrid Mamba--Attention--MoE architectures, and compute importance scores for MoE experts, embedding channels, Mamba heads and head channels, attention heads and FFNs.

\textbf{Embedding channels:} We aggregate normalized input activations across sequence ($L$) and batch ($B$) dimensions to compute the importance score for each embedding channel $i$ as $\text{F}^{\text{emb}}_i = \sum_{B,L} |\text{LN}(X)|_i$.

\textbf{FFN and MoE channels:} Importance score for channel $i$ is computed from the activations of the first up-projection layer, summed over batch and sequence length as $\text{F}^{\text{ffn}}_i = \sum_{B,L} |X(W_{up})^T|_i$.

\textbf{Mamba heads and channels:} Following the group-aware constraints that preserve SSM structure introduced in~\cite{taghibakhshi2025minitron} and used in Nemotron Elastic~\citep{taghibakhshi2025nemotron}, we compute head channel scores based on the Mamba input projection layer activations as $F^{head\_channel}_d = \left\|\sum_{B,L} s_{:,d}\right\|_2$, where $s = \text{LN}(X)(W_d)^T$. These scores are then aggregated across all heads. Individual head scores are computed using top-ranked channel activations ($\mathcal{D}_{\text{top}}$), as $F^{head}_h = \|s_{h,\mathcal{D}_{\text{top}}}\|_2$, where $h \in \{1,\ldots,m_h\}$ and $m_h$ denotes the number of mamba heads.

\textbf{Attention heads:} Importance for each head is computed from the attention scores aggregated across batch and sequence length as $\text{F}^{\text{attn}}_h = \sum_{B,L} \|Attn(XW^Q_h, XW^K_h, XW^V_h)\|_2$.

\textbf{MoE expert importance (REAP):} For MoE layers with conditional expert activation, we adopt Router-Weighted Expert Activation Pruning (REAP)~\citep{lasby2025reap}, which measures each expert's direct contribution to the layer output magnitude. The saliency score for expert $f_j$ is computed as $S_j = \frac{1}{|\mathcal{X}_j|} \sum_{x \in \mathcal{X}_j} g_j(x) \cdot \|f_j(x)\|_2$, where $\mathcal{X}_j$ is the set of inputs where $g_j(x) \in \text{TopK}(g(x))$, and $g_j(x)$ is the router gate-value for expert $j$. This criterion approximates expert importance by averaging the product of routing weights and expert output magnitudes over tokens where the expert is active. Unlike naive frequency-based pruning, REAP considers both routing frequency and functional contribution, enabling principled expert selection for heterogeneous MoE elastification.

\paragraph{Depth.} Layer importance is estimated iteratively using normalized mean squared error (MSE) between the full model's logits and those obtained with individual layers removed. At each iteration, for every remaining layer $j$, we compute $s_j = \frac{\sum_{B,L}(\mathcal{M}_{\text{full}} - \mathcal{M}_{-j})^2}{\sum_{B,L}\mathcal{M}^2_{\text{full}}}$, where $\mathcal{M}_{\text{full}}$ denotes logits from the full model and $\mathcal{M}_{-j}$ denotes logits with layer $j$ removed in addition to previously pruned layers. The iterative procedure accounts for changes in layer importance conditioned on previously pruned layers~\cite{muralidharan2024compact,taghibakhshi2025minitron,taghibakhshi2025nemotron}. Normalization ensures that scores are comparable across calibration datasets.

\subsection{Elastic Formulation}

We describe our nested, weight-shared architecture, which allows a single hybrid Mamba--Attention--MoE LLM to adapt dynamically to varying resource constraints. The model can be resized along both width and depth, enabling instant extraction of sub-networks with different parameter budgets and active expert counts without additional fine-tuning.

\paragraph{Elastic Width.}
We define elastic choices for each axis and sub-network $\mathcal{S}$: embedding dimension $d_e^\mathcal{S}$, attention heads $n_h^\mathcal{S}$, Mamba heads $m_h^\mathcal{S}$ and head channels $m_d^\mathcal{S}$, MoE expert count $e^\mathcal{S}(\ell)$, and FFN intermediate dimension $f^\mathcal{S}(\delta)$. Sub-networks are constructed by selecting values from these axes according to a target budget. The FFN and MoE components use layer-specific indices, with $1\le\delta\le N_{\text{F}}$ and $1\le\ell\le N_{\text{MoE}}$, since their widths can be chosen independently per layer to support heterogeneous configurations, where \(N_F\) and \(N_{\text{MoE}}\) denote the number of FFN and MoE layers, respectively.
In all cases, elastic choices form nested hierarchies: for each axis, we select a subset of components ranked by importance, such that smaller-budget sub-networks $\mathcal{S}$ always use the most salient contiguous components retained by larger-budget variants.

\paragraph{Elastic Depth.}
For the depth axis, elasticity is controlled by a binary selection vector $\gamma^\mathcal{S} = [\gamma^\mathcal{S}_0, \gamma^\mathcal{S}_1, \ldots, \gamma^\mathcal{S}_{N-1}]$, where $\gamma^\mathcal{S}_i \in \{0,1\}$ indicates whether layer $i$ is active in sub-network $\mathcal{S}$. The choice of which layer to disable is determined by its importance score, with less important layers removed first.

\paragraph{Implementation.}
Full implementation details, including elastification of each axis, masking, and zero-shot slicing, are provided in Appendix~\ref{app:implementation}.

\subsection{Elastic Training}

\paragraph{Router architecture and design:}
We train dedicated routers to learn the optimal model-specific mapping between the user-provided budget and the corresponding value for each nested axis. Each router consists of two fully connected layers with leaky ReLU activation applied between them.

The input to the router, for each axis, is a one-hot encoded vector representing the target compression level (budget specification): $u^{(\texttt{axis})} = \zeta_l \in \mathbb{R}^{n_{\text{targets}}}$, where $\texttt{axis}\in(d_e, m_h, m_d, n_h, e, f)$, $\zeta_l$ is the $l$-th standard basis vector, and $n_{\text{targets}}$ is the number of target model configurations. For example, with three budgets (100\%, 70\%, 50\%) as illustrated in Figure~\ref{fig:overview}, the inputs are $u = [1,0,0]$, $[0,1,0]$, and $[0,0,1]$ respectively.

Each router is parameterized as: $h^{(\texttt{axis})} = \text{LeakyReLU}(W^{(\texttt{axis})}_1 u^{(\texttt{axis})} + b^{(\texttt{axis})}_1)$  followed by $z^{(\texttt{axis})} = W^{(\texttt{axis})}_2 h^{(\texttt{axis})} + b^{(\texttt{axis})}_2$
where $W^{(\texttt{axis})}_1 \in \mathbb{R}^{d_{\text{router}} \times n_{\text{targets}}}$ and $W^{(\texttt{axis})}_2 \in \mathbb{R}^{n^{(\texttt{axis})}_{\text{out}} \times d_{\text{router}}}$ are learnable parameters.

For heterogeneous configurations, the router outputs target size per layer: $n^{(\texttt{moe\_expert, het})}_{\text{out}} = |\mathcal{E}| \times N_{\text{MoE}}$; $n^{(\texttt{ffn, het})}_{\text{out}} = |\mathcal{F}| \times N_F$, where $|\mathcal{E}|$ and $|\mathcal{F}|$ denote the cardinality of target expert counts and FFN dimension sets.

\paragraph{Loss formulation:}
The router outputs are passed through Gumbel-Softmax with temperature $\tau$ to produce soft probability distributions $P^{(\texttt{axis})}_i = \frac{\exp\left(\frac{\kappa\text{log}\pi^{(\texttt{axis})}_i + g_i}{\tau}\right)}{\sum_j \exp\left(\frac{\kappa\text{log}\pi^{(\texttt{axis})}_j + g_j}{\tau}\right)}$ over configuration choices for each elastic axis, where $g_i \sim \text{Gumbel}(0,1)$;  $\kappa$ is the scaling factor to balance the relative magnitude of logits; $\tau$ is a temperature parameter that controls the smoothness of the approximation, and as $\tau \rightarrow 0$, the $P^{(\texttt{axis})}$ distribution approaches a one-hot vector.

At each training iteration, we sample from these distributions to obtain relaxed discrete selections that enable gradient flow to the router parameters. The router is trained to optimize a resource-aware objective that maps selected configurations to hardware and computational constraints: $\mathcal{L}_{\text{router}} = \|\mathcal{C} - \hat{\mathcal{C}}\|$, where $\mathcal{C}$ is the resource cost of configuration (parameter count, memory usage, latency, or throughput) chosen by the router and $\hat{\mathcal{C}}$ is the target resource cost.

\paragraph{Knowledge distillation and multi-budget optimization:}
Following Minitron~\cite{muralidharan2024compact}, \NOne{} leverages knowledge distillation~\cite{hinton2015distilling} exclusively during elastic training, using the non-elastified parent model as the teacher to guide both architecture choice and accuracy optimization through teacher-aligned signals. We compute $\mathcal{L}_{\text{KD}} = D_{\text{KL}}(p_\varphi(x;\tau) \| p_\theta(x;\tau))$,
where $p_\varphi(x;\tau)$ denotes the teacher's softmax output at temperature $\tau$, and $p_\theta(x;\tau)$ denotes the student elastic model's corresponding distribution.

The final objective combines $\mathcal{L}_{\text{KD}}$ with a router based loss:
$\mathcal{L}_{\text{total}} = \mathcal{L}_{\text{KD}}(\theta) + \lambda \cdot \mathcal{L}_{\text{router}}(\psi)$, where $\psi$ denotes router parameters, and $\lambda > 0$ balances model accuracy against resource constraints.

This end-to-end optimization enables the router to make architecture choices directly from the actual training signal, rather than relying on zero-shot proxy metrics evaluated post-hoc---a key distinction from prior methods that decouple architecture configuration from training objectives. Our approach prioritizes sub-networks with strong capacity for continued learning under knowledge distillation, rather than those that simply minimize loss at initialization.

\paragraph{Two-stage training with curriculum-based sampling:}
Multi-budget elastic training (i.e., jointly training multiple nested models) requires carefully orchestrated data allocation across budget targets to prevent training imbalance and maintain accuracy across all sub-networks. Empirically, naïve uniform sampling in extended-context regimes causes the full-budget model to degrade while smaller budgets improve, indicating gradient competition and motivating our curriculum-based sampling strategy~\citep{taghibakhshi2025nemotron} (see also Appendix~\ref{app:budeg-sampling-ablation}). Hence, we introduce a two-stage training pipeline that adapts the sampling strategy to context length:

\textit{Stage 1: Uniform Budget Sampling (Short Context).}
During the initial short-context phase (sequence length $L_1 \approx 8192$, total tokens $T_1$), we employ uniform budget sampling. For $n_b$ target budgets, each training batch receives equal allocation with probability $\frac{1}{n_b}$. Uniform sampling ensures the router receives balanced training signals from all sub-networks, allowing architecture exploration without budget-specific bias.

\textit{Stage 2: Curriculum-Based Non-Uniform Sampling (Extended Context).}
During extended-context training (sequence length $L_2 = 49152$, total tokens $T_2$), we transition to non-uniform sampling that prioritizes full-budget models:
$p_2(b) = \alpha_b, \quad \forall b \in \{1,\ldots,n_b\}$, where $\sum_{i=1}^{n_b} \alpha_i = 1$ and weights are typically skewed toward larger budgets.

\paragraph{The role of extended-context training for reasoning:}
Reasoning tasks require extended token budgets for multi-step inference, making short-context elastic training insufficient for developing true reasoning capability. Training with 49K-token contexts exposes elastic variants to long reasoning chains, motivating our two-stage curriculum that first recovers accuracy for smaller models in the family and then incorporates reasoning-specific abilities (refer to Appendix~\ref{app:two-stage-ablation} for details, and to~\cite{taghibakhshi2025nemotron} for analogous Nano v2 ablations).

\section{Elastic Budget Control}
\label{sec:method:budget}

Existing budget control methods, such as the ones used in the NVIDIA Nemotron Nano models~\citep{nano2025efficient, blakeman2025nemotron3}, work in two phases:
(1) \textbf{thinking phase}: generation starts by prompting the model to produce reasoning text via a pre-pended \texttt{<think>} token. The number of generated thinking tokens is monitored and the phase is terminated once the predefined budget (e.g., $2k$ tokens) is reached;
(2) \textbf{answering phase}: the model is then forced to generate the final answer conditioned on the (potentially incomplete) reasoning text.

Unfortunately, existing budget control frameworks are forced to rely on static model architectures; i.e., they use the exact same model across prefill, thinking and answering. This results in computational inefficiencies when uniform capacity is applied to phases of heterogeneous complexity. In practice, reasoning and answer synthesis often exhibit disparate computational demands. This discrepancy motivates {\em elastic budget control}: a mechanism to dynamically calibrate model capacity for each inference phase, aligning resource allocation with task requirements to achieve the optimal performance--efficiency trade-off.

Specifically, using two model variants of differing sizes---denoted as $\mathcal{M}_{L}$ (Large) and $\mathcal{M}_{S}$ (Small)---we evaluate four experimental scenarios:
\begin{itemize}[nosep]
    \item $\textbf{$\mathcal{M}_{L} \to \mathcal{M}_{L}$}$: $\mathcal{M}_{L}$ used for both phases.
    \item $\textbf{$\mathcal{M}_{S} \to \mathcal{M}_{S}$}$: $\mathcal{M}_{S}$ used for both phases.
    \item $\textbf{$\mathcal{M}_{L} \to \mathcal{M}_{S}$}$: $\mathcal{M}_{L}$ for thinking, $\mathcal{M}_{S}$ for answering.
    \item $\textbf{$\mathcal{M}_{S} \to \mathcal{M}_{L}$}$: $\mathcal{M}_{S}$ for thinking, $\mathcal{M}_{L}$ for answering.
\end{itemize}

Guided by the empirical findings detailed in Section~\ref{sec:budget_control_exp}, we identify \textbf{$\mathcal{M}_{S} \to \mathcal{M}_{L}$} as the optimal configuration for elastic budget control. This design choice is rooted in the asymmetric computational requirements of the two phases: (1) \textbf{high-volume reasoning}: the thinking phase benefits from a larger token budget to explore complex reasoning paths. By utilizing $\mathcal{M}_{S}$, we can generate extensive reasoning traces with minimal computational overhead; (2) \textbf{high-fidelity synthesis}: the answering phase requires superior instruction-following and internal consistency to transform the reasoning trace into a correct final response. $\mathcal{M}_{L}$ provides the necessary cognitive capacity to ensure this synthesis is robust.

\paragraph{Cache state sharing:}
In our elastic budget control experiments, we align with the standard inference practice of recomputing cache states when switching between nested models. This strategy is consistent with recent approaches adopted in the Nemotron Nano models~\citep{nano2025efficient, blakeman2025nemotron3} and Qwen3~\citep{yang2025qwen3}, reflecting the current constraints of inference frameworks like vLLM and TensorRT-LLM.

While naive budget control (using a single fixed model) naturally allows for cache reuse, switching between a smaller ``thinking'' model and a larger ``answering'' model typically breaks cache compatibility. \NOne{} mitigates this issue by preserving the structure of Mamba and attention layers during elastification, maintaining cache compatibility across nested models. Empirically, we find cache states to be highly consistent across nested models, indicating that cache states can be safely transplanted between nested models (see Tables~\ref{tab:cache_sim} and ~\ref{tab:transplant_gsm8k} in Appendix~\ref{app:cache}). As a result, although our reported wall-clock times currently include cache recomputation overhead, our architecture enables seamless cache reuse once framework support becomes available. Consequently, our reported performance should be viewed as a conservative lower bound, with further gains expected as cache reuse is enabled.

\section{Experiments and Results}
\label{sec:experiments}

We apply our \NOne{} method to Nemotron Nano v3 30B/3.6A~\cite{blakeman2025nemotron3}, a hybrid Mamba--Transformer--MoE reasoning model. Analogous experiments on the Mamba--Attention-only Nemotron Nano v2 12B parent (yielding 9B and 6B nested variants from a 110B-token elastic run) are reported in the original Nemotron Elastic paper~\citep{taghibakhshi2025nemotron}; we therefore refer the reader to that work for Nano v2 results, and focus the experiments here on the new MoE setting.

\subsection{Experimental Setup}

\paragraph{Nested compression:}
For Nano v3, we simultaneously train three nested models from a single parent architecture using multi-budget elastic compression: this includes the original parent model, and two smaller models. The router optimizes for target active parameter budgets of 3.6B, 2.8B, and 2.0B, deriving 23B (2.8A) and 12B (2.0A) nested variants from the 30B (3.6A) parent. The frozen version of the parent model (prior to elastification) serves as the teacher for knowledge distillation. Detailed architecture specifications for the router-selected configurations are provided in Appendix~\ref{app:architecture-details}.

\paragraph{Dataset:}
For importance estimation and knowledge distillation, we use the open-source data used to train the Nano v3 parent model~\cite{blakeman2025nemotron3}. Ablation studies on data blends are provided in Appendix~\ref{app:data-blend}.

\paragraph{Evaluation benchmarks:}
We evaluate across a comprehensive suite of reasoning and knowledge benchmarks: MMLU-Pro~\cite{wang2024mmlupro} (college-level multiple-choice reasoning), GPQA~\cite{rein2023gpqa} (graduate-level science questions), AIME-2024 and AIME-2025~\cite{aime} (invitational mathematics), LiveCodeBench v5~\cite{jain2024livecodebench} (code generation), IFBench~\cite{pyatkin2025generalizing} (instruction following), and Tau Bench~\cite{barres2025tau} (industry-specific reasoning across airline, retail, and telecom domains).

\paragraph{Hyperparameters and training setup:}
For importance estimation, we use 1024 calibration samples with sequences of length 8192. Knowledge distillation is performed in two phases: \NOne{} on Nano v3 is trained for $\sim$100B tokens (batch 6144, sequence length 8k), followed by $\sim$60B tokens (batch 1024, sequence length 49k), totaling 160B tokens. The full elastic training pipeline is integrated into NVIDIA Megatron-LM\footnote{\href{https://github.com/NVIDIA/Megatron-LM}{https://github.com/NVIDIA/Megatron-LM}} and run on the same training stack used for the parent Nano v3 model, which makes scaling \NOne{} to larger parents and longer contexts straightforward.

\textbf{Optimizer settings:}
Model parameters are trained with a learning rate of $10^{-4}$, while the router uses $10^{-2}$. A 60-step linear warmup is applied to all parameters. The Gumbel-Softmax temperature $\tau$ is annealed from 1.0 to 0.05. The router loss coefficient $\lambda$ is set to 1.0, and $\kappa$, the linear scaling coefficient for router logits, is scaled linearly from 1.0 to 10.0. The router intermediate hidden dimension ($d_{\text{router}}$) is set to 256.

\paragraph{Budget sampling strategy:}
During the short-context phase, we employ uniform budget sampling with $p(\text{budget}) = 1/3$ for each model variant. In the extended-context phase, we transition to weighted non-uniform sampling: $p(\text{30B}) = 0.5$, $p(\text{23B}) = 0.3$, $p(\text{12B}) = 0.2$. This prevents accuracy degradation with the full model during extended-context training. Ablations and details on the budget sampling strategy follow~\cite{taghibakhshi2025nemotron}, with additional Nano v3 specifics in Appendix~\ref{app:budeg-sampling-ablation}.

\subsection{Main Results}

Detailed accuracy results for the elastic Nano v3 variants are shown in Table~\ref{tab:nanov3-results}, with average scores in Figure~\ref{fig:side_by_side_elastic} (Left). \NOne{} Elastic-30B matches the accuracy of its parent model, while smaller nested variants remain highly competitive against similarly-sized community models. Two-stage training with adjusted budget sampling prevents accuracy degradation in larger models (see~\cite{taghibakhshi2025nemotron} and Appendix~\ref{app:budeg-sampling-ablation}). Overall, \NOne{} scales effectively to MoE-based hybrid architectures, producing elastic models comparable to or surpassing independently trained counterparts with a fraction of the training budget.

\begin{table*}[tb]
\centering
\small
\setlength{\tabcolsep}{4pt}
\renewcommand{\arraystretch}{1.05}
\caption{Detailed \NOne{} (Nano v3) results. All three Elastic variants are obtained from a single 160B-token training run. \textbf{*} Indicates the teacher model used for distillation.}
\resizebox{\textwidth}{!}{%
\begin{tabular}{l|ccc|cc}
\hline
\textbf{Benchmark} & \textbf{NanoV3 Elastic-12B} & \textbf{NanoV3 Elastic-23B} & \textbf{NanoV3 Elastic-30B} & \textbf{NanoV3-30B*} & \textbf{Qwen3-30B-A3B} \\
 & \textbf{(2.0A)} & \textbf{(2.8A)} & \textbf{(3.6A)} & \textbf{(3.6A)} & \textbf{(3.3A)} \\
\hline
AIME-2025 & 78.54 & 85.63 & \textbf{88.54} & 87.92 & 80.00 \\
GPQA & 57.39 & 69.82 & 72.10 & \textbf{73.11} & 70.83 \\
LiveCodeBench v5 & 55.24 & 67.30 & \textbf{72.70} & 71.75 & 68.25 \\
MMLU-Pro & 68.28 & 76.07 & 78.63 & 78.86 & \textbf{81.11} \\
IFBench (prompt) & 64.03 & 67.43 & \textbf{70.58} & 70.82 & 43.28 \\
IFBench (instruct) & 67.39 & 70.75 & \textbf{73.96} & 73.19 & 46.57 \\
Tau-Airline & 24.67 & 38.67 & 43.33 & 44.67 & \textbf{52.67} \\
Tau-Retail & 49.12 & 55.56 & \textbf{59.36} & 53.51 & 56.43 \\
Tau-Telecom & 29.33 & 30.99 & \textbf{33.33} & 30.99 & 28.36 \\
\hline
\end{tabular}}
\label{tab:nanov3-results}
\end{table*}

\subsection{Elastic Budget Control Results}
\label{sec:budget_control_exp}

\begin{figure}[tb]
\centering
\includegraphics[width=1.0\linewidth]{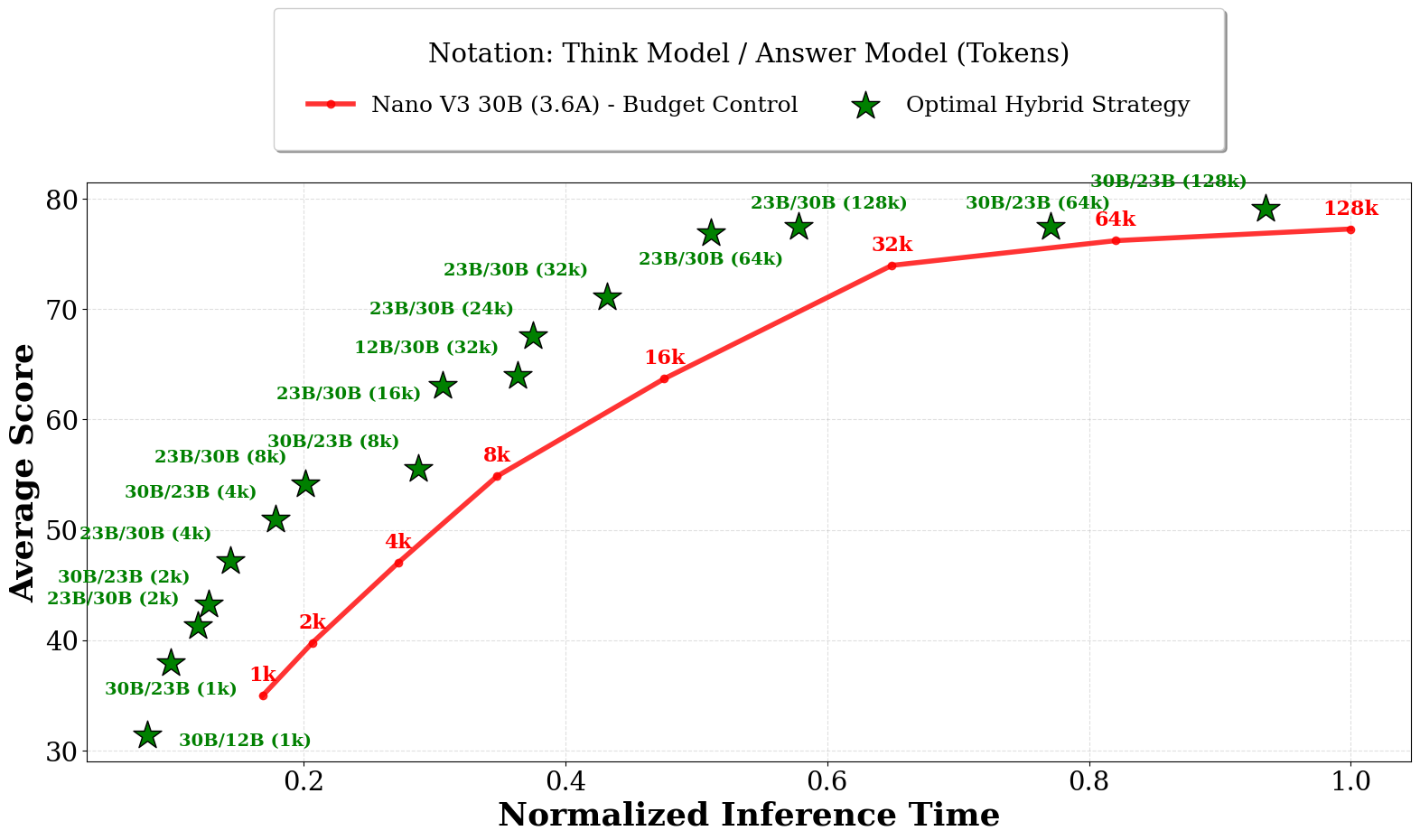}
\caption{Elastic budget control configurations across computational budgets for Nemotron Nano v3. The $\mathcal{M}_{S} \to \mathcal{M}_{L}$ strategy generally offers the best accuracy--latency tradeoffs ($23\text{B} \to 30\text{B}$), while $30\text{B} \to 23\text{B}$ is preferred at the highest latency regime. $12\text{B} \to 30\text{B}$ / $30\text{B} \to 12\text{B}$ provide low-latency options at the smallest budgets (refer to Appendix~\ref{app:budget-control} for more details).}
\label{fig:method_comparison}
\end{figure}

As described in Section~\ref{sec:method:budget}, \NOne{} enables the use of different nested models during distinct reasoning stages such as thinking and answer generation.
Figure~\ref{fig:side_by_side_elastic} (right) shows the results for our proposed elastic budget control method and compares it to the default Nano v3 budget control. We notice from the figure that our approach achieves accuracy improvements up to 16\% and latency reductions of up to 1.9$\times$.
Figure~\ref{fig:method_comparison} summarizes the elastic budget control configurations underlying the accuracy--speed Pareto improvements in Figure~\ref{fig:side_by_side_elastic} (right). We compare hybrid thinking--answering allocations across models and observe that the $\mathcal{M}_{S} \to \mathcal{M}_{L}$ strategy (small thinking, large answering), specifically in the case of $23\text{B} \to 30\text{B}$, provides the best accuracy--latency tradeoffs over a wide range of budgets, since it concentrates capacity on the final answer while keeping the reasoning phase cheaper. For applications that require the highest absolute accuracy and can afford more inference time, the $30\text{B} \to 23\text{B}$ configuration is preferred, slightly outperforming other settings at the high-latency end of the frontier. At the lowest budgets, either $12\text{B} \to 30\text{B}$ or $30\text{B} \to 12\text{B}$ can be used to minimize latency while still benefiting from asymmetric thinking--answering allocations.
To measure inference latency, we utilized the NeMo-Skills library\footnote{\href{https://github.com/NVIDIA-NeMo/Skills}{https://github.com/NVIDIA-NeMo/Skills}} with vLLM as the backend, employing BF16 precision. We set the batch sizes to 40, 30, and 16 for the 12B, 23B, and 30B models, respectively. Detailed per-benchmark Pareto curves and configuration-level analyses, including a corresponding scatter plot for all combinations of thinking and answering, are provided in Appendix~\ref{app:budget-control}.

\subsection{Runtime Speedups}
Table~\ref{tab:speedups} shows the throughput improvements that \NOne{} (Nano v3 Elastic) models achieve at an input and output sequence length of 8192 and 16384, respectively; we use the maximum batch size that fits each model on an H100 GPU at bfloat16 precision. We notice that \NOne{} nested models achieve significant speedups (up to 2.4$\times$ in this case) while also enabling inference at much higher batch sizes on the same GPU. Across-precision and across-GPU throughput numbers (BF16/FP8/NVFP4 on RTX 5080/5090/Pro 6000) are reported in Section~\ref{sec:quantization} (Figure~\ref{fig:rtx-tput}).

\begin{table}[tb]
\centering
\small
\setlength{\tabcolsep}{3pt}
\renewcommand{\arraystretch}{1.05}
\caption{Throughput improvements for \NOne{} (Nano v3) models. Throughput is measured with vLLM at bfloat16 on an H100 GPU.}
\resizebox{\linewidth}{!}{%
\begin{tabular}{l|c|c|c}
\hline
\textbf{Elastic Variant} & \textbf{ISL/OSL} & \textbf{Max Batch} & \textbf{Speedup} \\
\hline
Nano v3 30BA3.6A & 8192/16384 & 36 & 1$\times$ \\
Nano v3 23BA2.8A & 8192/16384 & 108 & 1.8$\times$ \\
Nano v3 12BA2A & 8192/16384 & 224 & 2.4$\times$ \\
\hline
\end{tabular}}
\label{tab:speedups}
\end{table}

\subsection{Cost Efficiency Analysis}

\paragraph{Training efficiency:}
A key advantage of \NOne{} (and Nemotron Elastic more broadly~\citep{taghibakhshi2025nemotron}) is that it eliminates the exploratory training runs required by prior methods such as Minitron~\cite{muralidharan2024compact} and Minitron-SSM~\cite{taghibakhshi2025minitron}, which prune and distill multiple candidate architectures to select the best for final knowledge distillation, incurring token costs that scale linearly with the number of models. In contrast, \NOne{} performs end-to-end, router-guided architecture selection in a single elastic training run, simultaneously optimizing all target budgets. Nested training further provides regularization from the full model, reducing the token requirement even for the final distillation stage. Quantitative token-budget comparisons for deriving 6B/9B variants from the Nano v2 12B parent, as well as the corresponding training-token and deployment-memory scaling analyses, are reported in~\cite{taghibakhshi2025nemotron}; for Nano v3, our 160B-token elastic run analogously delivers all three nested variants in a single pass, in contrast to $n \cdot (\text{Tokens}_{\text{explore}} + \text{Tokens}_{\text{KD}})$ scaling for prior compression methods.

\paragraph{Deployment memory efficiency:}
Elastic models with nested weight-sharing offer substantial memory savings: all variants share a single parameter space, with only lightweight routing metadata distinguishing them. Once training is complete, architecture selection can be hard-coded, enabling zero-shot slicing with no runtime overhead. Deploying multiple nested models requires memory equal to the largest model, $\text{Memory}_{\text{Nested}}(n) = \text{Size}(\text{Model}_{\max})$, compared to linear scaling for separate checkpoints, $\text{Memory}_{\text{Separate}}(n) = \sum_{i=1}^n \text{Size}(\text{Model}_i)$. For example, elastic Nano v3 (12B, 23B, 30B) fits in 58.9~GB BF16 versus 126.1~GB for separate variants, highlighting the memory efficiency of \NOne{}'s nested models (Table~\ref{tab:memory-comparison}). The analogous Nano v2 comparison (24~GB vs.~54~GB) is reported in~\cite{taghibakhshi2025nemotron}.

\begin{table}[t!]
\centering
\setlength{\tabcolsep}{12pt}
\renewcommand{\arraystretch}{1.05}
\caption{Deployment memory comparison for Nemotron Nano v3. Elastic models host multiple budgets in a single nested checkpoint, using less than half the memory of storing separate checkpoints for each variant\protect\footnotemark.}
\resizebox{\linewidth}{!}{%
\begin{tabular}{l|c|c}
\hline
\textbf{Config} & \textbf{Models} & \textbf{Mem (BF16)} \\
\hline
Elastic (V3) & 12B+23B+30B & \textbf{58.9~GB} \\
NanoV3 & 12B+23B+30B & 126.1~GB \\
\hline
\end{tabular}}
\label{tab:memory-comparison}
\end{table}
\footnotetext{Nano v3 23B2.8A and Nano v3 12B2A in Table~\ref{tab:memory-comparison} are estimated based on scaling laws and have not been trained as standalone checkpoints.}

\section{Quantized Elastic Models}
\label{sec:quantization}

A practical bottleneck for deploying large reasoning models is the memory and bandwidth cost of carrying high-precision weights, even after elastification. While BF16 \NOne{} checkpoints already provide a $\sim$2.2$\times$ memory reduction over storing 30B/23B/12B variants separately, modern GPUs can additionally exploit narrower precisions such as FP8 (E4M3) and NVFP4 to further compress weights and activations. A naive route would be to quantize each elastic variant independently after slicing; however, this would (i) require a separate quantization run per variant, (ii) break the nested weight-sharing property---since post-hoc per-variant quantization no longer preserves a single shared parameter space across budgets---and (iii) prevent zero-shot extraction of smaller variants from a single quantized checkpoint.

Instead, we preserve a \emph{single nested checkpoint} under quantization. We release two quantized elastic checkpoints for Nano v3, both containing the full 30B/23B/12B nested family and supporting the same zero-shot slicing pipeline as the BF16 model.

\paragraph{Overview of quantization formats:}
We use two formats, both defined per output channel of each linear layer:
\begin{itemize}[nosep]
    \item \textbf{FP8 (E4M3):} Per-tensor scaling with E4M3 mantissa-exponent format. Weights are stored in FP8; KV/SSM states are kept in BF16.
    \item \textbf{NVFP4:} NVIDIA's 4-bit floating-point format with two-level scaling factors (E4M3 per-block scales and FP32 per-tensor scale). Weights are stored in NVFP4. The quantization configuration is the same one that was used for Nemotron-3-Nano-NVFP4~\cite{xin2026qad} i.e. we keep the 6 self-attention layers and their preceding Mamba-2 layers in BF16 precision, quantize the remaining network to NVFP4, and quantize KV-Cache to FP8. 
\end{itemize}

\paragraph{Quantization Approach:}
We start from the BF16 elastic checkpoint and apply quantization while preserving the nested structure. For FP8, post-training quantization (PTQ) is sufficient: we quantize the BF16 checkpoint and then apply the nested masks to obtain all variants. This preserves the shared parameter space and yields strong accuracy without retraining.
For NVFP4, PTQ leads to substantial degradation (4.12\% average accuracy drop on the 30B variant). To recover this loss, we apply nested Quantization-Aware Distillation (QAD) on top of the quantized model.

\paragraph{Nested QAD:}
Given a trained elastic BF16 teacher $\mathcal{T}$ with parameters $\Theta_{\max}^{\text{bf16}}$ and the same router-determined nested mask hierarchy used at deployment, we initialize a quantized student $\mathcal{S}_q$ with the same architecture and router masks but quantized weights $\Theta^q$ (NVFP4). At each training step, we sample a budget $b \in \{30\text{B}, 23\text{B}, 12\text{B}\}$ following the same curriculum-based, non-uniform sampling distribution used during BF16 elastification (Section~\ref{sec:methodology}), apply the corresponding budget-conditioned masks to only the student $\mathcal{S}_q$, and minimize a KL divergence between teacher and student logits:
\begin{equation}
\mathcal{L}_{\text{QAD}} = \mathbb{E}_{b \sim p_2}\big[D_{\text{KL}}\big(p_{\mathcal{T}}(\cdot \mid x) \,\big\|\, p_{\mathcal{S}_q}(\cdot \mid x; b)\big)\big],
\end{equation}
where $p_2$ is the same Stage-2 budget distribution used in BF16 elastification. Because the student is trained with the same nested mask hierarchy while being distilled from a non-elastified teacher, gradients from the smaller budgets regularize the most salient shared weights, while gradients from the full budget shape the tail; this follows the same multi-budget optimization intuition that justifies BF16 elastic training, transposed to the quantized regime.

\paragraph{Nested QAD recipe:}
We reuse the Stage-2 extended-context KD dataset and run a short distillation phase with a context length of 48K. We use a batch size of 512, and keep the learning rate to a small constant value of $1\mathrm{e}{-5}$. We train for 200 steps ($\sim$5B tokens) and select the checkpoint with the best accuracy on the downstream benchmarks across the elastic variants.

\paragraph{Accuracy recovery:}
Aggregated across the benchmark suite of Section~\ref{sec:experiments}, both FP8 and NVFP4 nested variants recover the vast majority of BF16 accuracy (Table~\ref{tab:quant-recovery}) while preserving the nested structure, i.e., the 23B and 12B variants are obtained via zero-shot slicing from the corresponding FP8/NVFP4 30B checkpoint.

  \begin{table}[t]
  \centering
  \small
  \setlength{\tabcolsep}{4pt}
  \renewcommand{\arraystretch}{1.05}
  \caption{Nested-quantized accuracy recovery for \NOne{}-Nano v3. Recovery is computed as the ratio of the average
  benchmark accuracy of the quantized variant to that of the corresponding BF16 elastic variant.}
  \begin{tabular}{l|cc}
  \hline
  \textbf{Variant} & \textbf{FP8} & \textbf{NVFP4} \\
  \hline
  30B (3.6A) & 98.69\% & 97.79\% \\
  23B (2.8A) & 99.03\% & 99.15\% \\
  12B (2.0A) & 100.26\% & 97.10\% \\
  \hline
  \end{tabular}
  \label{tab:quant-recovery}
  \end{table}

\paragraph{Per-benchmark FP8 and NVFP4 results:}
Per-benchmark BF16, FP8, and NVFP4 numbers across the three nested variants on the full evaluation suite (including MATH-500) are reported in Table~\ref{tab:quant-results} of Appendix~\ref{app:quant-perbench}.

\begin{figure}[t]
\centering
\includegraphics[width=\linewidth]{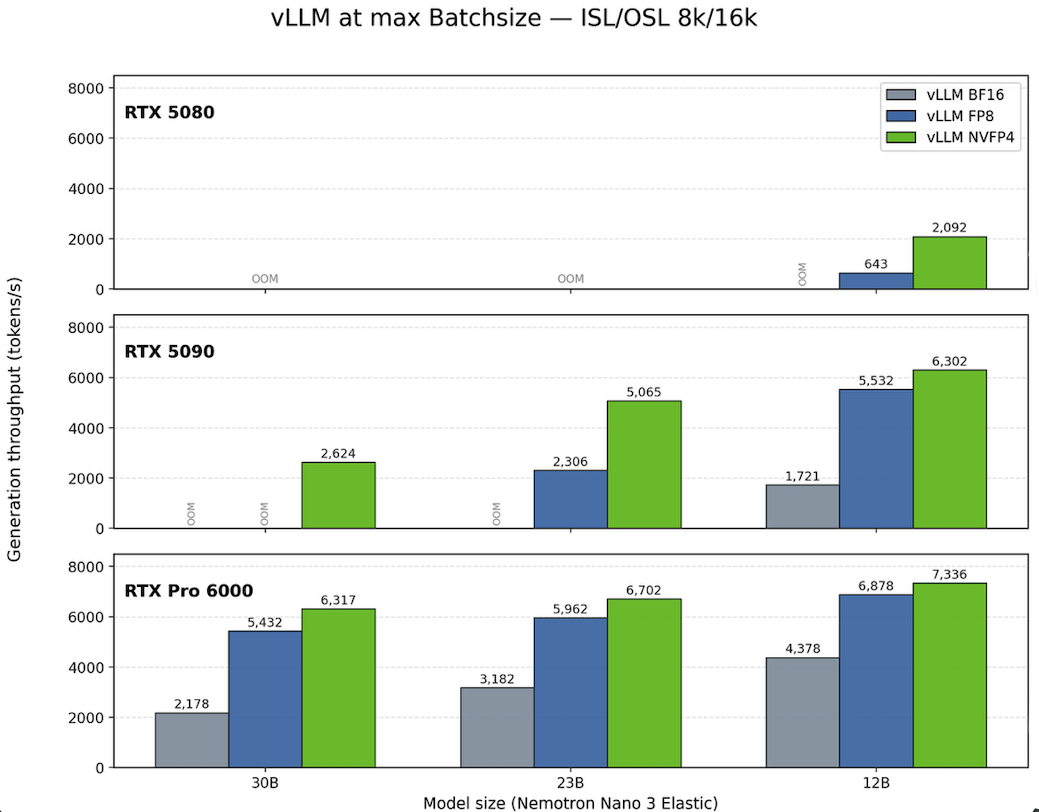}
\caption{Generation throughput (tokens/s) of \NOne{} (Nano v3) variants under vLLM at max batch size with ISL/OSL = 8k/16k, on RTX 5080 (top), RTX 5090 (middle), and RTX Pro 6000 (bottom), in BF16, FP8 (PTQ), and NVFP4 (QAD). ``OOM'' marks configurations that do not fit. NVFP4 enables deployment of larger nested budgets on smaller GPUs and dominates throughput at every operating point.}
\label{fig:rtx-tput}
\end{figure}

\paragraph{Throughput and deployment memory:}
Quantized elastic checkpoints offer further deployment savings on top of the nested BF16 baseline. We summarize device memory for loading BF16/FP8/NVFP4 variants of the nested 12B/23B/30B elastic family in Table~\ref{tab:quant-memory}, measured under vLLM on RTX Pro 6000, RTX 5090, and RTX 5080 (reported memory is device-agnostic at load time in our setup). At ISL$=8192$ and OSL$=16384$, Figure~\ref{fig:rtx-tput} reports maximum aggregate decode throughput (tokens/s) for each variant on those GPUs; entries marked OOM did not complete successfully in our sweep. The combination of nested elasticity and quantization opens up deployment regimes that are unreachable with the BF16 30B parent: the 12B NVFP4 variant fits and runs on an RTX 5080 (where every BF16 budget OOMs), and on an RTX 5090 the 30B/23B variants are only deployable in NVFP4. On the higher-memory RTX Pro 6000, every variant is deployable at every precision and NVFP4 still delivers the best throughput, with the 12B NVFP4 model reaching 7{,}426~tokens/s---a 3.4$\times$ improvement over the 30B BF16 baseline (2{,}178~tokens/s).

\begin{table}[t]                                                                                                        
  \centering                                                                                                              
  \small                                                                                                                  
  \setlength{\tabcolsep}{6pt}                                                                                             
  \renewcommand{\arraystretch}{1.05}                                                                                      
  \caption{Per-variant device memory (GB) to load each nested-elastic \NOne{} (Nano v3) standalone. Because the variants
  are nested, deploying a family of any size requires storing only the largest checkpoint.}                               
  \begin{tabular}{l|ccc}
  \hline                                                                                                                  
  \textbf{Variant} & \textbf{BF16} & \textbf{FP8} & \textbf{NVFP4} \\           
  \hline                                                                                                                  
  30B (3.6A) & 58.9 & 31.4 & 18.7 \\                                                                                      
  23B (2.8A) & 44.0 & 23.7 & 14.1 \\
  12B (2.0A) & 23.2 & 13.0 & 8.0 \\                                                                                       
  \hline                                                                                                                  
  \end{tabular}
  \label{tab:quant-memory}                                                                                                
  \end{table}   

\paragraph{Take-aways:}
Quantized elastification preserves the single-checkpoint paradigm while adapting to format-specific constraints. FP8 supports direct PTQ on the BF16 elastic checkpoint with negligible accuracy loss, whereas NVFP4 requires a short nested QAD phase to recover the degradation introduced by PTQ. In both cases, the nested structure is maintained, enabling zero-shot extraction of 30B/23B/12B variants from a single quantized checkpoint. This retains the deployment flexibility of \NOne{} while further reducing memory and improving efficiency across the accuracy–latency trade-off.

\section{Related Work}
\label{sec:related}

\textbf{Hybrid SSM--Transformer--MoE Models.} \NOne{} focuses on hybrid SSM--Transformer--MoE and SSM--Transformer architectures, both of which have demonstrated strong performance for general tasks and efficient long-context modeling~\citep{gu2023mamba,dao2024transformers,lieber2024jamba,glorioso2024zamba,blakeman2025nemotron,blakeman2025nemotron3}. Concurrent compression work~\citep{taghibakhshi2025minitron} introduces group-aware Mamba pruning but requires separate distillation per model size; \NOne{} sidesteps this by adopting nested elastification across the full Mamba--Attention--MoE design space.

\textbf{Elastic and Nested Architectures.} MatFormer~\citep{kudugunta2023matformer} and Flextron~\citep{cai2024flextron} pioneered nested weight-sharing for Transformers. MatMamba~\citep{shukla2024matmambamatryoshkastatespace} introduces Matryoshka-style sub-block architecture for Mamba layers. FlexGS~\citep{Liu_2025_CVPR} extends this methodology into the computer vision domain. Most directly, Nemotron Elastic~\citep{taghibakhshi2025nemotron} introduces the first elastic post-training framework for hybrid Mamba--Attention reasoning LLMs (Nano v2), with a learnable router, two-stage curriculum, and zero-shot slicing. \NOne{} builds directly on Nemotron Elastic and extends it along three orthogonal axes: (1) elastification of Mamba--Attention--MoE parents (Nano v3), with REAP-based MoE expert ranking and heterogeneous per-layer expert/FFN selection; (2) \emph{elastic budget control}, which leverages the nested family at inference to allocate different submodels to the thinking and answering phases; and (3) elastification in the quantized regime via QAD, producing nested NVFP4 / FP8 checkpoints. To the best of our knowledge, no prior work jointly addresses these three settings in a single elastic checkpoint.

\textbf{Reasoning Model Efficiency.} Reasoning-capable LLMs generate extended thought chains for complex problem-solving~\citep{wei2022chain,yao2023tree}. While prior work focuses on improving reasoning model efficiency via prompting strategies or reinforcement learning~\citep{lightman2023lets}, we explore a novel axis (model size, plus phase-wise model selection) to achieve the same objective.

\textbf{Quantization for LLMs.} Post-training quantization (PTQ) and quantization-aware training (QAT) have been widely studied for dense Transformer LLMs, typically targeting INT8/FP8 weights and activations. Recent work extends these recipes to FP4-class formats. \NOne{} differs in that it performs \emph{quantization-aware distillation on top of an elastic teacher}: the same nested mask hierarchy is applied to both teacher and student, so quantization preserves rather than collapses the elastic structure, and a single quantized checkpoint supports zero-shot slicing across the full model family.

\section*{Conclusions}

We present \NOne{}, the first post-training method for producing elastic, reasoning-capable hybrid Mamba--Transformer--MoE LLMs that simultaneously unlocks efficient many-in-one model families and elastic budget control at inference. Building on the Nemotron Elastic framework~\citep{taghibakhshi2025nemotron}, \NOne{} efficiently derives 23B and 12B model variants from Nemotron Nano v3 30B using only $\sim$160B training tokens, achieving a $360\times$ reduction over training from scratch and a $7\times$ reduction versus sequential compression. Additionally, elastic budget control at inference improves the accuracy--latency trade-off, achieving up to 16\% higher accuracy and $1.9\times$ lower latency. Through post-training quantization and quantization-aware distillation, we further produce FP8 and NVFP4 nested checkpoints that preserve zero-shot slicing while substantially shrinking the deployment footprint. All nested models share a constant memory footprint, enabling efficient deployment. Overall, our approach enables state-of-the-art accuracy with adaptive inference and makes training and deploying a family of reasoning models highly cost-effective.

\textbf{Limitations and future work.} Our current approach achieves $\sim$2.5$\times$ compression for the smallest variant; exploring extreme compression ratios (e.g., 10$\times$) for ultra-resource-constrained settings remains an open challenge. Further, task-specific elastic routing (i.e., automatically selecting optimal model configurations based on the input domain, such as code, math, or multilingual tasks) requires further study. Cache-state transplantation across nested variants is another promising direction whose practical realization is currently bottlenecked by inference framework support.

\section{Acknowledgments}
We would like to thank our colleagues and leaders at NVIDIA for their valuable input and support, including Alex Fit-Florea, Bill Dally, Bryan Catanzaro, Joey Conway, Jonah Alben, Jonathan Cohen, Kari Briski, Luis Vega, Michael Lightstone, Nave Assaf, Oleksii Kuchaiev, Terry Kong, Udi Karpas.

\clearpage
{
  \small
  \bibliographystyle{unsrt}
  \bibliography{paper}
}

\appendix
\section{Router-Selected Architecture Specifications}
\label{app:architecture-details}

\subsection{Nemotron Nano v3}

The router optimizes for target \textit{active parameter} budgets. Due to the limitation of vLLM (which is used for all benchmark evaluations) in supporting heterogeneous MoE FFN channels, the router is set to select homogeneous configurations across layers.

All variants share 32 attention heads, 64 Mamba heads, 128 MoE experts and the layer pattern. The embedding dimensions and MoE FFN dimensions for each budget are shown in Table~\ref{tab:nanov3-arch}.

\begin{table}[h]
\centering
\small
\setlength{\tabcolsep}{5pt}
\caption{\NOne{} (Nano v3) architecture variants.}
\resizebox{\linewidth}{!}{%
\begin{tabular}{l|ccc}
\hline
& \textbf{30B (3.6A)} & \textbf{23B (2.8A)} & \textbf{12B (2.0A)} \\
\hline
Embedding Dim & 2688 & 2304 & 1920 \\
MoE FFN Dim & 1856 & 1600 & 960 \\
\hline
\end{tabular}}
\label{tab:nanov3-arch}
\end{table}

\noindent\textbf{Layer Pattern:} M-E-M-E-M*-E-M-E-M-E-M*-E-M-E-M-E-M*-E-M-E-M-E-M*-E-M-E-M-E-M*-E-M-E-M-E-M-E-M-E (M = Mamba, E = MoE, * = Attention).

For analogous Nano v2 architecture specifications, see~\cite{taghibakhshi2025nemotron}.

\section{Ablation Studies}

To validate key design choices, we summarize the ablations performed for the \NOne{}-Nano v3 setting. The corresponding Nano v2 ablations are reported in detail in~\cite{taghibakhshi2025nemotron} and the conclusions transfer directly.

\subsection{Effects of Two-Stage Training}
\label{app:two-stage-ablation}
\begin{table*}[h]
\centering
\small
\setlength{\tabcolsep}{6pt}
\renewcommand{\arraystretch}{1.1}
\caption{Two-stage training improvements (Nano v2). Stage 2 (extended-context) provides substantial gains on reasoning benchmarks, particularly AIME-2025, where smaller models benefit significantly (6B: +19.8\%, 9B: +9.7\%). Reproduced from~\cite{taghibakhshi2025nemotron} for completeness, since this is the empirical anchor that motivates our two-stage curriculum on Nano v3 as well.}
\resizebox{\textwidth}{!}{%
\begin{tabular}{l|cc|c|c}
\hline
\textbf{Model (Benchmark)} & \textbf{Stage 1} & \textbf{Stage 2} & \textbf{Absolute Gain} & \textbf{Relative Improvement} \\
\hline
NanoV2 Elastic-6B (MATH-500) & 95.15 & 96.50 & +1.35 & +1.4\% \\
NanoV2 Elastic-6B (AIME-2025) & 56.88 & 68.13 & +11.25 & +19.8\% \\
NanoV2 Elastic-6B (GPQA) & 49.12 & 53.78 & +4.66 & +9.5\% \\
\hline
NanoV2 Elastic-9B (MATH-500) & 97.13 & 97.25 & +0.12 & +0.1\% \\
NanoV2 Elastic-9B (AIME-2025) & 68.75 & 75.42 & +6.67 & +9.7\% \\
NanoV2 Elastic-9B (GPQA) & 59.43 & 62.50 & +3.07 & +5.2\% \\
\hline
NanoV2 Elastic-12B (MATH-500) & 97.27 & 97.70 & +0.43 & +0.4\% \\
NanoV2 Elastic-12B (AIME-2025) & 72.92 & 75.83 & +2.91 & +4.0\% \\
NanoV2 Elastic-12B (GPQA) & 62.50 & 63.25 & +0.75 & +1.2\% \\
\hline
\end{tabular}}
\label{tab:two-stage-efficacy}
\end{table*}

Table~\ref{tab:two-stage-efficacy} reveals that Stage 2 extended-context training delivers significant improvements on complex reasoning benchmarks, especially for smaller models. The 6B model gains 19.8\% on AIME-2025, while the 12B model gains 4.0\%, indicating that smaller models particularly benefit from extended-context adaptation for multi-step reasoning. We adopt the same two-stage curriculum for Nano v3 elastification.

\subsection{Impact of Budget Sampling Strategy}
\label{app:budeg-sampling-ablation}

The budget sampling ablation comparing uniform vs.~adjusted Stage-2 sampling is reported in detail in~\cite{taghibakhshi2025nemotron}. Adjusted sampling ($p(\text{12B}) = 0.5, p(\text{9B}) = 0.3, p(\text{6B}) = 0.2$ for Nano v2; analogous weighting for Nano v3) substantially improves performance for the full-budget model on challenging reasoning benchmarks (e.g., +3.54 percentage points on AIME-2025 for the Nano v2 12B model). For Nano v3, we apply the analogous distribution $p(\text{30B}) = 0.5, p(\text{23B}) = 0.3, p(\text{12B}) = 0.2$ in Stage 2 and observe the same qualitative behavior.

\subsection{Data Blend Ablation Study}
\label{app:data-blend}

To determine the optimal training data composition for elastic compression of Nano v3, we conduct a comprehensive ablation study across multiple dimensions: data blend ratio (reasoning vs. pretraining data), sequence length, prompt masking strategy, and inclusion of teacher-generated RL samples.

For Nemotron Nano v2, we adopt the data blend ratio established in prior compression studies~\cite{nano2025efficient}, which identified 70\% reasoning (post-training) + 30\% pretraining as optimal. We maintain this blend throughout the Nano v2 experiments reported in~\cite{taghibakhshi2025nemotron}. For Nemotron Nano v3, no compression study exists in the accompanying report~\cite{blakeman2025nemotron3}. We therefore begin with the same 70\% reasoning + 30\% pretraining blend as our baseline and systematically ablate across additional dimensions: 100\% reasoning variants, prompt masking strategies, sequence lengths, and augmenting the dataset with RL rollouts from the parent model, as detailed below.

All ablations are performed on a 15\% embedding-pruned student model derived from Nemotron Nano v3 30B. The same calibration data is used for both importance estimation and knowledge distillation across each configuration.

\subsubsection{Ablation Dimensions}

We systematically vary four key dimensions:

\begin{itemize}
\item \textbf{Data blend ratio:} 100\% reasoning (post-training) vs. 70\% reasoning + 30\% pretraining.
\item \textbf{Sequence length:} 8K vs. 32K vs. 49K vs. 256K tokens.
\item \textbf{Prompt masking:} With prompt masking vs. without prompt masking.
\item \textbf{RL augmentation:} With and without teacher-generated RL rollouts.
\end{itemize}

\subsubsection{Stage 1: Initial Configuration Selection (15B Tokens)}

Table~\ref{tab:data-blend-stage1} presents results for the first stage of ablation, where all configurations are trained for 15B tokens at their respective sequence lengths.

\begin{table*}[h]
\centering
\setlength{\tabcolsep}{5pt}
\renewcommand{\arraystretch}{1.2}
\caption{Stage 1 data blend ablation (15B tokens). The configuration with 70\% reasoning + 30\% PT has the best overall average. The effect of sequence length is negligible among 8K, 32K, and 49K, while the performance drops significantly at 256K. Data blend abbreviation: PT = Pretraining, SFT = Supervised Finetuning Reasoning, RL = Reinforcement Learning rollouts.}
\resizebox{\linewidth}{!}{
\begin{tabular}{cl|ccccccc|c}
\hline
& \textbf{Configuration (15B token training)} & \textbf{MATH} & \textbf{AIME} & \textbf{GPQA} & \textbf{LiveCode} & \textbf{MMLU} & \textbf{IFB-P} & \textbf{IFB-I} & \textbf{Avg} \\
\hline
\multicolumn{10}{l}{\textit{Pruning-only baseline (0B tokens, no training)}} \\
1a & 100\% reasoning, mask, 0B & 89.50 & 40.83 & 40.09 & 38.10 & 59.41 & 48.43 & 53.07 & 52.78 \\
1b & 70\% SFT + 30\% PT, no mask, 0B & 89.90 & 40.83 & 41.60 & 38.41 & 59.22 & 44.55 & 47.17 & 51.67 \\
1c & 100\% reasoning, no mask, 0B & 89.05 & 39.77 & 44.57 & 41.27 & 59.61 & 46.73 & 50.81 & 53.12 \\
\hline
\multicolumn{10}{l}{\textit{Masking ablation}} \\
2a & 100\% reasoning, mask, 8K & 96.75 & 82.29 & 63.63 & 63.17 & 74.98 & 53.88 & 56.96 & 70.24 \\
\textbf{2c} & \textbf{100\% reasoning, no mask, 8K} & 97.20 & 85.42 & 68.31 & 66.03 & 77.04 & 55.91 & 57.67 & \textbf{72.51} \\
\hline
\multicolumn{10}{l}{\textit{Blend ratio ablation (70\% vs. 100\% reasoning)}} \\
\textbf{3a} & \textbf{70\% SFT + 30\% PT, no mask, 8K} & 97.30 & 85.11 & 69.19 & 65.71 & 76.94 & 59.01 & 60.60 & \textbf{73.41} \\
3b & 100\% reasoning, no mask, 8K & 97.20 & 85.42 & 68.31 & 66.03 & 77.04 & 55.91 & 57.67 & 72.51 \\
\hline
\multicolumn{10}{l}{\textit{Extended sequence length (32K, 49K, 256K)}} \\
4a & 70\% SFT + 30\% PT, no mask, 32K & 97.20 & 88.13 & 69.44 & 66.35 & 77.55 & 55.98 & 58.45 & 73.30 \\
\textbf{4b} & \textbf{70\% SFT + 30\% PT, no mask, 49K} & 97.50 & 86.46 & 67.99 & 65.07 & 76.90 & 58.84 & 62.03 & \textbf{73.54} \\
4c & 70\% SFT + 30\% PT, no mask, 256K & 96.75 & 84.58 & 66.41 & 61.26 & 76.58 & 56.46 & 59.58 & 71.66 \\
\hline
\multicolumn{10}{l}{\textit{RL augmentation ablation}} \\
5a & 30\% RL + 49\% SFT + 21\% PT, 8K & 97.65 & 84.17 & 67.55 & 63.81 & 76.91 & 59.31 & 61.61 & 73.00 \\
5b & 50\% RL + 35\% SFT + 15\% PT, 8K & 97.50 & 85.21 & 69.19 & 66.66 & 76.93 & 57.14 & 59.58 & 73.17 \\
5c & 70\% RL + 21\% SFT + 9\% PT, 8K & 96.95 & 83.75 & 66.47 & 66.35 & 76.72 & 57.48 & 60.77 & 72.64 \\
\hline
\end{tabular}}
\label{tab:data-blend-stage1}
\end{table*}

\subsubsection{Stage 2: Context Extension Strategy (Additional 10B Tokens)}

After identifying the optimal Stage 1 configuration (70\% reasoning + 30\% pretraining, 8K, no prompt mask, no RL rollouts), we investigate context extension strategies by training for an additional 10B tokens (total 25B). We compare three approaches: (1) continuing at 8K throughout, (2) extending from 8K to 49K in Stage 2, and (3) training at 49K from the start. Results are shown in Table~\ref{tab:data-blend-stage2}.

\begin{table*}[h]
\centering
\setlength{\tabcolsep}{5pt}
\renewcommand{\arraystretch}{1.2}
\caption{Stage 2 context extension ablation (25B total tokens). Each configuration starts from the corresponding Stage 1 checkpoint (Table~\ref{tab:data-blend-stage1}). Extending from 8K to 49K (74.40\%) outperforms both 8K-only (72.37\%) and 49K-from-start (73.70\%) strategies.}
\resizebox{\linewidth}{!}{
\begin{tabular}{cl|ccccccc|c}
\hline
& \textbf{Configuration (extra 10B token training)} & \textbf{MATH} & \textbf{AIME} & \textbf{GPQA} & \textbf{LiveCode} & \textbf{MMLU} & \textbf{IFB-P} & \textbf{IFB-I} & \textbf{Avg} \\
\hline
\multicolumn{10}{l}{\textit{8K only (no context extension)}} \\
3a$\rightarrow$ & 8K continued, same blend & 97.05 & 82.92 & 67.42 & 64.76 & 77.17 & 57.48 & 59.76 & 72.37 \\
\hline
\multicolumn{10}{l}{\textit{8K $\rightarrow$ 49K context extension}} \\
\textbf{3a}$\rightarrow$ & \textbf{49K extension, same blend} & 97.40 & 87.08 & 69.94 & 67.94 & 77.49 & 59.25 & 61.67 & \textbf{74.40} \\
3a$\rightarrow$ & 49K extension, 50\% RL + 35\% SFT + 15\% PT & 97.50 & 86.67 & 68.81 & 66.67 & 77.47 & 59.66 & 62.87 & 74.24 \\
5b$\rightarrow$ & 49K extension, 50\% RL + 35\% SFT + 15\% PT & 97.20 & 88.33 & 69.26 & 66.35 & 77.15 & 57.35 & 61.79 & 73.92 \\
\hline
\multicolumn{10}{l}{\textit{49K from start (no staged extension)}} \\
4b$\rightarrow$ & 49K continued, same blend & 97.65 & 87.50 & 68.75 & 65.71 & 77.63 & 58.02 & 60.65 & 73.70 \\
4b$\rightarrow$ & 49K continued, 100\% RL & 97.20 & 86.04 & 68.18 & 64.36 & 76.96 & 59.93 & 61.91 & 73.51 \\
4b$\rightarrow$ & 49K continued, 50\% RL + 35\% SFT + 15\% PT & 97.30 & 84.58 & 68.11 & 67.94 & 76.85 & 57.82 & 60.53 & 73.30 \\
\hline
\end{tabular}}
\label{tab:data-blend-stage2}
\end{table*}

\subsubsection{Key Findings}

\paragraph{Stage 1 findings:}
\begin{enumerate}
\item \textbf{Sequence length:} 8K achieves the best balance between computational efficiency and performance. While longer contexts (32K, 49K) provide modest gains on specific benchmarks (e.g., AIME-2025), they do not consistently improve average performance and incur significantly higher training costs.

\item \textbf{Prompt masking:} Removing prompt masking consistently improves performance by 1.5--2.5 percentage points across all benchmarks, suggesting that allowing the model to attend to prompts during training improves reasoning capability.

\item \textbf{Blend ratio:} The 70\% reasoning + 30\% pretraining blend outperforms 100\% reasoning by 0.9 percentage points on average. The inclusion of pretraining data provides beneficial diversity and prevents overfitting to post-training distributions.

\item \textbf{RL augmentation:} Adding teacher-generated RL rollouts does not provide consistent improvements and occasionally degrades performance on reasoning-heavy benchmarks, potentially due to distribution mismatch.
\end{enumerate}

\paragraph{Stage 2 findings (context extension):}
\begin{enumerate}
\item \textbf{Staged context extension is superior:} The 8K$\rightarrow$49K extension strategy (74.40\% average) outperforms both 8K-only (72.37\%) and 49K-from-start (73.70\%) approaches, achieving the best balance between recovery and long-context capability.

\item \textbf{Recovery before extension:} Training at 49K from the beginning yields slightly lower performance than staged extension despite using 2B more tokens (27B vs. 25B). This suggests that the compressed model benefits from initial recovery at shorter contexts before adapting to extended sequences.

\end{enumerate}

\subsubsection{Final Configuration}

Based on these findings, we adopt the following two-stage training recipe for all elastic compression experiments:

\textbf{Stage 1 (short-context):} 70\% reasoning + 30\% pretraining, 8K sequence length, no prompt masking

\textbf{Stage 2 (extended-context):} 70\% reasoning + 30\% pretraining, 49K sequence length, no prompt masking

For Nano v3 we use a total budget of 160B tokens (100B + 60B). The corresponding 110B-token configuration for Nano v2 is reported in~\cite{taghibakhshi2025nemotron}.

\subsection{Ablation on Depth vs.~Width Compression}
\label{app:depth-vs-width}

As elastic compression can target both depth (layer count) and width (hidden dimensions, expert count) of neural networks, we compare the two approaches in isolation early in the study.

\subsubsection{Depth vs.~Width Compression Investigation}

We conducted preliminary ablations on elastic depth compression (15\% layer reduction) and width compression (15\% reduction across experts, FFN hidden dimensions, and embedding dimensions) on Nemotron Nano v3 base model. Both configurations were trained with 25B tokens of knowledge distillation from the full Nano v3 30B base (pretrained only) model as teacher.

Results in Table~\ref{tab:width-compression} reveal important tradeoffs between the two approaches. Width-based compression achieves 98.1\% of baseline performance (0.768 vs. 0.783 average), maintaining competitive accuracy across nearly all benchmarks. In contrast, depth compression achieves 95.2\% of baseline performance (0.745 vs. 0.783 average), with noticeable degradation on HumanEval and MMLU-Pro ($\sim$5.4\% each). Given the superior performance of width-based approaches, we prioritize width-based elastic compression for the main results, though depth compression remains viable for extreme latency-constrained scenarios.

\begin{table*}[htbp]
\centering
\small
\setlength{\tabcolsep}{3pt}
\renewcommand{\arraystretch}{1.05}
\caption{Depth vs.~width compression comparison with 15\% parameter reduction and 25B token knowledge distillation. \NOne{} width compression achieves 98.1\% of baseline performance (0.768 vs. 0.783 average). Elastic depth compression achieves 95.2\% of baseline (0.745 vs. 0.783 average).}
\resizebox{\textwidth}{!}{%
\begin{tabular}{l|ccccccccc|c}
\hline
\textbf{Config} & \textbf{ARC} & \textbf{GSM8K} & \textbf{HumanEval} & \textbf{MATH} & \textbf{MBPP} & \textbf{MMLU-Pro} & \textbf{MMLU} & \textbf{RACE} & \textbf{Wino} & \textbf{Avg} \\
\hline
Baseline & 0.918 & 0.915 & 0.703 & 0.775 & 0.724 & 0.613 & 0.782 & 0.884 & 0.762 & 0.783 \\
Depth (15\%) & 0.883 & 0.892 & 0.649 & 0.728 & 0.666 & 0.559 & 0.727 & 0.868 & 0.737 & 0.745 \\
Width (15\%) & 0.908 & 0.886 & 0.696 & 0.724 & 0.711 & 0.604 & 0.764 & 0.874 & 0.744 & 0.768 \\
\hline
\end{tabular}}
\label{tab:width-compression}
\end{table*}

\subsubsection{Practical Implications}

For most deployment scenarios prioritizing accuracy, we recommend width-based elastic compression as implemented in \NOne{}. However, for extreme latency-constrained scenarios where inference speed is paramount, and some accuracy loss is acceptable, depth-based compression via layer skipping can be easily activated as an optional mechanism. \NOne{}, as described in Section~\ref{sec:methodology} and Appendix~\ref{app:implementation}, supports the layer skipping mechanism.

However, depth reduction is omitted in the final models to explore potential cache reuse benefits (see Appendix~\ref{app:cache}).

\section{Per-Benchmark Elastic Budget Control Results}
\label{app:budget-control}

This appendix provides detailed per-benchmark analysis of the elastic budget control results presented in Figure~\ref{fig:side_by_side_elastic} (right). We evaluate hybrid elastic budget control configurations on Nemotron Nano v3 across six benchmarks: AIME-2025, GPQA, MMLU-Pro, LiveCodeBench v5, and IFBench (prompt-strict and instruction-strict modes).

The evaluated configurations are:
\begin{itemize}[nosep]
    \item Baseline: 30B model for both thinking and answering
    \item 23B-30B: 23B for thinking, 30B for answering
    \item 30B-23B: 30B for thinking, 23B for answering
    \item 12B-30B: 12B for thinking, 30B for answering
    \item 30B-12B: 30B for thinking, 12B for answering
\end{itemize}

\subsection{Overall Performance and Pareto Frontier}

\begin{figure*}[htbp]
\centering
\includegraphics[width=\linewidth]{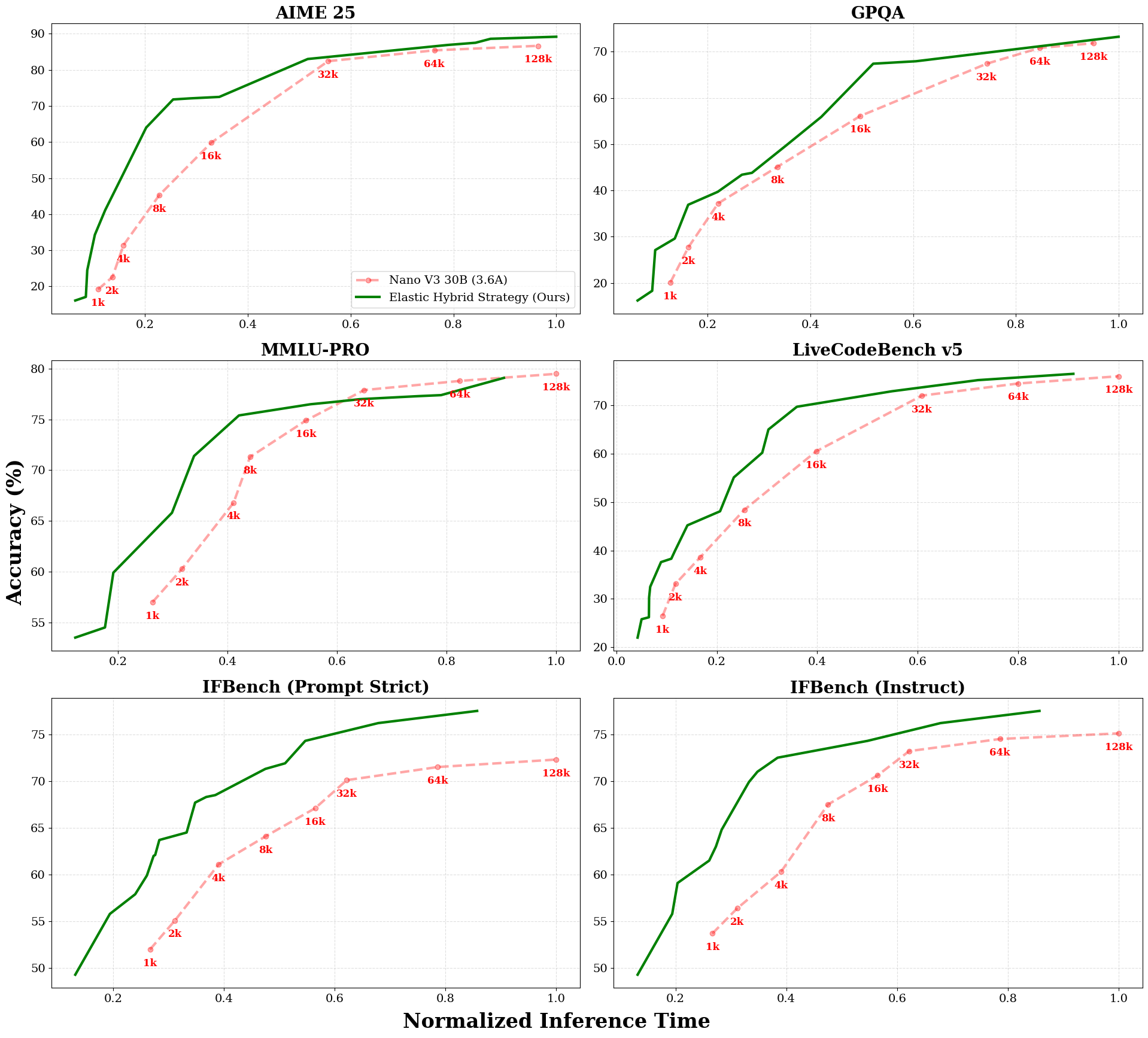}
\caption{Per-benchmark Pareto frontiers comparing hybrid elastic budget control (colored curves) against vanilla Nano v3 budget control (gray). Across all benchmarks, elastic configurations dominate the Pareto frontier, achieving superior accuracy--speed tradeoffs.}
\label{fig:budget-pareto-all}
\end{figure*}

Figure~\ref{fig:budget-pareto-all} demonstrates that hybrid elastic budget control consistently improves the accuracy vs. speed Pareto frontier compared to vanilla Nemotron Nano v3 budget control across all benchmarks. The elastic configurations achieve higher accuracy at equivalent inference times or faster inference at equivalent accuracy, validating the effectiveness of phase-specific model selection.

\subsection{Configuration-Specific Analysis}

\begin{figure}[htbp]
\centering
\includegraphics[width=1.0\linewidth]{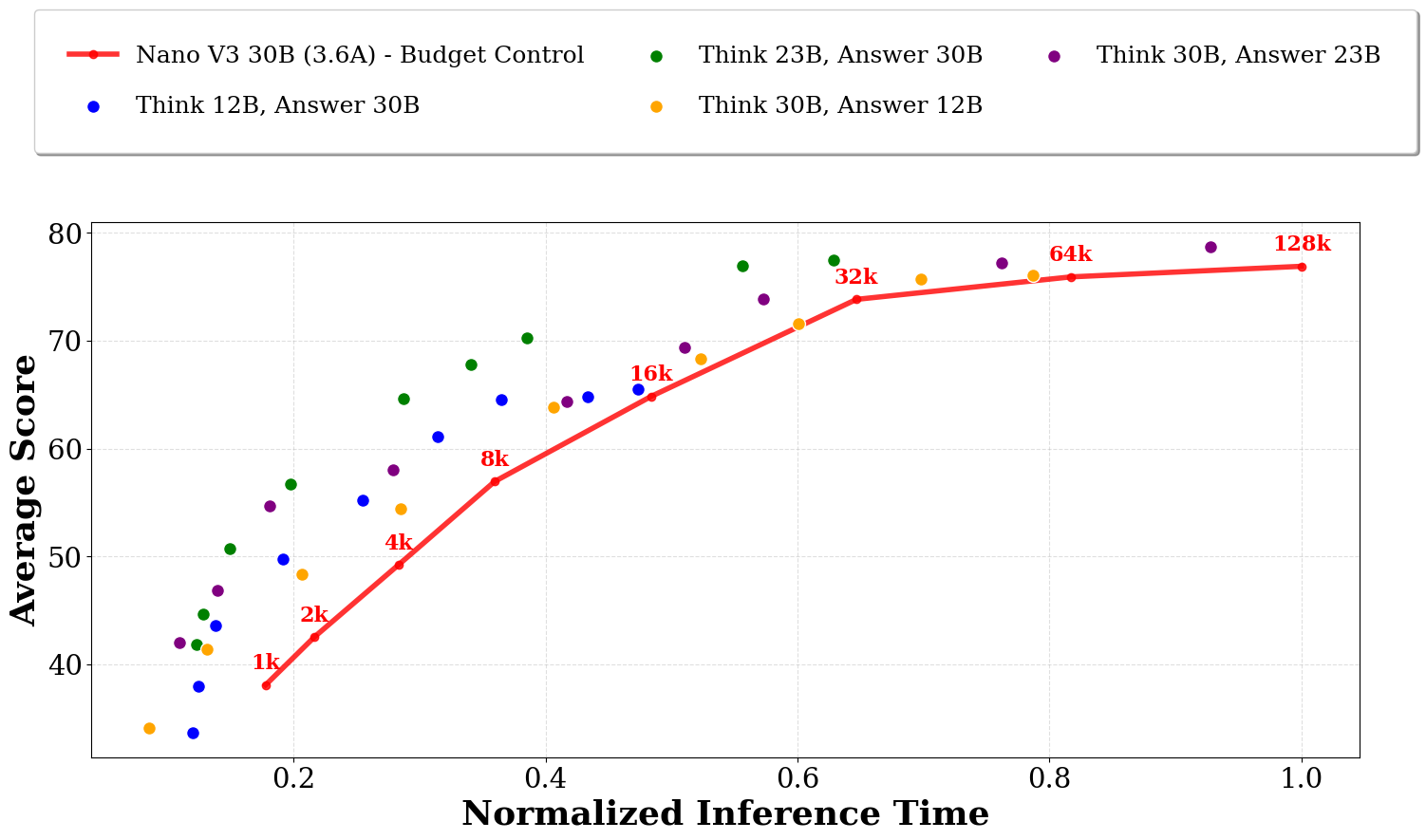}
\caption{Detailed configuration comparison across accuracy--speed space. The 23B-30B configuration dominates at higher scores and moderate inference times, while 12B-30B and 30B-12B are optimal for extremely low-latency scenarios.}
\label{fig:budget-scatter-all}
\end{figure}

Figure~\ref{fig:budget-scatter-all} provides a detailed breakdown of which configurations are optimal at different operating points. The 23B-30B configuration (23B thinking, 30B answering) achieves the best overall balance, particularly excelling with higher accuracy and moderate inference times. For extremely low-latency scenarios where inference time is minimized, the 12B-30B and 30B-12B configurations provide optimal performance, while high latency scenarios prefer the 30B-23B configuration.

\subsection{Budget Control Length Stats}
In this section, we analyze the token distribution across three distinct components: prompt, thinking process, and final answer. Table \ref{tab:length_stats} details the average token lengths for each benchmark under a fixed computation budget ($16k$). Note that the average thinking length remains below the budget cap because many samples naturally conclude their generation process before exhausting the full allocation.

\begin{table}[ht]
    \centering
    \caption{Comparison of budget control statistics. The table shows results for GPQA with a fixed computation budget of $16k$ across different Think--Answer setups.}
    \label{tab:length_stats}
    \resizebox{\linewidth}{!}{
    \begin{tabular}{l|c|c|c}
        \toprule
        Setup & Avg Prompt & Avg Reason & Avg Answer \\
        \midrule
        23B -- 30B & & 11020 & 1066 \\
        12B -- 12B & \multirow{7}{*}{257} & 12524 & 1291 \\
        12B -- 30B & & 12580 & 1270 \\
        23B -- 23B & & 10812 & 1062 \\
        30B -- 12B & & 10674 & 1106 \\
        30B -- 23B & & 10586 & 1033 \\
        30B -- 30B & & 10518 & 1035 \\
        \bottomrule
    \end{tabular}}
\end{table}

\section{Cache State Reuse}\label{app:cache}
As detailed in Section~\ref{sec:method:budget}, the nested structure of \NOne{} models preserves the parameters of Mamba and attention layers across different model depths. We hypothesize that this architectural consistency allows for the direct transplantation of KV and SSM state caches from a smaller ``thinking'' model (e.g., $\mathcal{M}_{S}$) to a larger ``answering'' model (e.g., $\mathcal{M}_{L}$) without the need for recomputation. In this section, we provide empirical evidence validating the structural consistency of these caches and quantify the impact of cache transplantation on generation quality. For all experiments in this section, we use the Nano v2 Elastic family from~\cite{taghibakhshi2025nemotron}.

To evaluate cache compatibility, we first compare the internal states (specifically, KV caches, Mamba SSM states, and Mamba convolutional states) generated by the nested sub-models against those of the full model, given an identical input sequence. We quantify this alignment by computing the average cosine similarity across shared layers. As shown in Table~\ref{tab:cache_sim}, we observe high cosine similarity across most state types, particularly for the Mamba SSM states (0.96) and Value caches (0.95), confirming strong structural alignment. While the convolutional states show lower similarity (0.71), their impact is limited due to their relatively small memory footprint (2.2 MB) compared to the dominant SSM states (70 MB). This suggests that the vast majority of the recurrent context is preserved during transplantation.

\begin{table}[htbp]
    \centering
    \renewcommand{\arraystretch}{1.1}
    \caption{Analysis of cache state compatibility between the nested sub-model and the full model given identical input sequences. Memory Footprint indicates the storage requirement for the cache states (BF16 precision). Note the distinct scaling behaviors: Key/Value caches grow linearly (adding 12 KB for every new token), whereas Conv/SSM states have a fixed total size ($O(1)$) that does not increase with sequence length. Cosine similarity is calculated independently for each shared layer and then averaged.}
    \label{tab:cache_sim}
    \resizebox{\linewidth}{!}{
    \begin{tabular}{l c c c c}
        \toprule
        State Type & Key Cache & Value Cache & Conv State & SSM State \\
        \midrule
        Mem. Footprint & 12 KB /tok. & 12 KB /tok. & 2.2 MB & 70 MB \\
        Cosine Sim. & 0.89 & 0.95 & 0.71 & 0.96 \\
        \bottomrule
    \end{tabular}}
\end{table}

Beyond structural similarity, we evaluate the impact of cache transplantation on downstream task performance using the GSM8k benchmark. Specifically, we use the compressed 6B sub-model to process the prompt and generate the cache states (KV, Conv, and SSM), which are then transplanted to the full 12B model for the answering phase.

Table~\ref{tab:transplant_gsm8k} presents the results. We observe that transplanting the full cache configuration (``All 6B cache'') results in a minimal accuracy drop compared to the standard 12B baseline (90.93\% vs. 90.18\%), a difference likely attributable to generation randomness rather than structural incompatibility. Interestingly, transplanting the convolutional states alone occasionally yields results slightly surpassing the baseline, suggesting that the local features captured by the smaller model are particularly robust. Overall, these results confirm that the transplanted states retain the critical reasoning information required for high-quality generation.

\begin{table}[htbp]
    \centering
    \caption{Ablation study of cache transplantation on GSM8k accuracy. The ``Baseline'' represents standard 12B inference. The ``Transplant'' rows denote experiments where the specific cache components are generated by the 6B sub-model and transferred to the 12B model for generation. The minimal performance drop in the ``All'' setting confirms the viability of cache reuse.}
    \label{tab:transplant_gsm8k}
    \begin{tabular}{l c}
        \toprule
        Method & Accuracy \\
        \midrule
        12B Model (Baseline) & 90.93\% \\
        \addlinespace
        Transplant 6B KV Cache & 89.26\% \\
        Transplant 6B Conv State & 91.34\% \\
        Transplant 6B SSM State & 89.24\% \\
        \addlinespace
        Transplant All 6B Cache & 90.18\% \\
        \bottomrule
    \end{tabular}
\end{table}

\section{Per-Benchmark Quantized Results}
\label{app:quant-perbench}

Table~\ref{tab:quant-results} provides the full per-benchmark breakdown of BF16, FP8 (PTQ), and NVFP4 (QAD) accuracy across the three nested \NOne{} (Nano v3) variants. AVG is the mean across MATH-500, AIME-2025, GPQA, LiveCodeBench v5, MMLU-Pro, IFBench (prompt and instruct), and the average of Tau-Airline/Retail/Telecom (denoted TAU).

\begin{table*}[t]
\centering
\small
\setlength{\tabcolsep}{4pt}
\renewcommand{\arraystretch}{1.05}
\caption{Per-benchmark results for BF16, FP8 (PTQ), and NVFP4 (QAD) across nested variants of \NOne{} (Nano v3).}
\resizebox{\textwidth}{!}{%
\begin{tabular}{l|ccccccccc}
\hline
\textbf{Model} & MATH-500 & AIME & GPQA & LCB & MMLU & IF-P & IF-I & TAU & AVG \\
\hline
30B BF16   & 97.70 & 88.54 & 72.10 & 72.70 & 78.63 & 70.58 & 73.96 & 45.34 & 74.94 \\
30B FP8    & 97.35 & 86.66 & 73.23 & 70.16 & 78.71 & 70.24 & 73.36 & 41.96 & 73.96 \\
30B NVFP4  & 97.60 & 86.46 & 72.60 & 68.25 & 78.17 & 69.18 & 72.24 & 41.80 & 73.29 \\
\hline
23B BF16   & 97.60 & 85.63 & 69.82 & 67.30 & 76.07 & 67.43 & 70.75 & 41.74 & 72.04 \\
23B FP8    & 97.35 & 83.33 & 68.18 & 66.59 & 76.05 & 66.33 & 69.85 & 43.07 & 71.34 \\
23B NVFP4  & 96.95 & 85.00 & 67.30 & 67.62 & 74.76 & 68.16 & 71.58 & 40.08 & 71.43 \\
\hline
12B BF16   & 96.00 & 78.54 & 57.39 & 55.24 & 68.28 & 64.03 & 67.39 & 34.37 & 65.16 \\
12B FP8    & 96.10 & 76.46 & 58.02 & 54.60 & 68.30 & 65.87 & 68.85 & 34.38 & 65.32 \\
12B NVFP4  & 95.85 & 74.17 & 55.43 & 54.29 & 62.49 & 64.42 & 67.88 & 31.62 & 63.27 \\
\hline
\end{tabular}}
\label{tab:quant-results}
\end{table*}

\section{Implementation Details}
\label{app:implementation}

\subsection{Implementation}

The elastic architecture is instantiated through structured masking applied to the hybrid Mamba--Attention--MoE and Mamba--Transformer models. Rather than modifying network topology or creating distinct sub-networks, we apply dimension-specific binary masks that dynamically select active components. This masking-based approach enables efficient training of multiple budgets simultaneously while maintaining architectural transparency and enabling straightforward deployment of any sub-network without architectural recompilation.

\subsubsection {Dynamic Model Formulation}

We present a flexible architecture framework for \NOne{} that enables dynamic adjustment of model dimensions during training through a structured masking approach. Our method builds upon the hybrid Mamba--Attention--MoE and Mamba--Transformer architectures and extends the elastic training paradigm to support comprehensive width and depth flexibility for hybrid architectures.

A dynamic model is obtained by making the stack of layers dynamic and then making each layer type dynamic across different dimensions. If the original LLM is defined as $y = \mathcal{L}_0^N(x)$ where $\mathcal{L}_0^j(x) = \mathcal{L}_0^{j-1}(x) + \mathcal{L}^j(\mathcal{L}_0^{j-1}(x))$, a dynamic layer stack is noted as $\mathcal{D} \circ \mathcal{L}_0^N$ where the operator $\mathcal{D}$ is applied to each layer and makes it dynamic. For example:
\begin{equation}
\mathcal{D} \circ \mathcal{L}^j = (\mathcal{D} \circ \mathcal{L}_j) \cdot \gamma_j
\end{equation}
where $\gamma_j \in \{0, 1\}$ controls layer retention (depth adaptation) and $\mathcal{D} \circ \mathcal{L}_j$ represents a dynamic Mamba, Attention, FFN, or MoE layer.

The dynamic operator $\mathcal{D}$ applies dimension-specific binary masks $\mathbf{m}$ to the output activations of each layer component, enabling selective feature retention (width adaptation):
\begin{equation}
\mathcal{D}(\mathcal{L}(x)) = \mathcal{L}(x) \odot \mathbf{m}
\end{equation}
where $\odot$ denotes element-wise multiplication and $\mathbf{m} \in \{0,1\}^d$ is a binary mask vector that determines which dimensions remain active. Depth adaptation is controlled through the binary coefficient vector $\boldsymbol{\gamma} = [\gamma_0, \gamma_1, \ldots, \gamma_{N-1}]$, while width adaptation is managed through dimension-specific masks applied within each layer type.

\subsubsection{Dynamic Mamba}

For Mamba-2 components in the hybrid architecture, we apply group-aware masking following permutation-preserving constraints to maintain structural integrity of state-space computations.

\paragraph{Dynamic Embedding Mask Operator.} The operator $\mathcal{M}_{\text{emb}}$ applies to any activation or weight matrix with the hidden size $d_e$ as one dimension. For a matrix $\mathbf{W} \in \mathbb{R}^{d_e \times k}$, the masked operation is:
\begin{equation}
\mathcal{M}_{\text{emb}}(\mathbf{W}) = \mathbf{W} \odot (\mathbf{I}_e \otimes \mathbf{1}_k)
\end{equation}
where $\mathbf{I}_e \in \{0,1\}^{d_e}$ with $\mathbf{I}_e[0:i] = 1$ and $\mathbf{I}_e[i+1:d_e] = 0$ for some $i \in [0, d_e]$, and $\otimes$ denotes outer product broadcasting across dimension $k$. For matrices $\mathbf{W} \in \mathbb{R}^{k \times d_e}$, the mask broadcasts similarly: $\mathcal{M}_{\text{emb}}(\mathbf{W}) = \mathbf{W} \odot (\mathbf{1}_k \otimes \mathbf{I}_e)$. This operator is applied to layer normalization outputs and all weight matrices interfacing with the embedding dimension.

\paragraph{Dynamic Mamba Mask Operator.} The operator $\mathcal{M}_{\text{mamba}}$ applies to matrices where dimensions derive from Mamba heads $m_h$ or head channels $m_d$. For a matrix $\mathbf{W} \in \mathbb{R}^{f(m_h, m_d) \times d_e}$ where $f$ represents a dimension function (typically $f(m_h, m_d) = m_h \cdot m_d$), the masked operation is:
\begin{equation}
\mathcal{M}_{\text{mamba}}(\mathbf{W}) = \mathbf{W} \odot (\mathbf{I}_m \otimes \mathbf{1}_{d_e})
\end{equation}
where $\mathbf{I}_m \in \{0,1\}^{f(m_h, m_d)}$ is constructed to satisfy:
\begin{equation}
\mathbf{I}_m[\phi(h,c)] = \begin{cases}
1 & \text{if } h \leq h^* \text{ and } c \leq c^* \\
0 & \text{otherwise}
\end{cases}
\end{equation}
with $\phi(h,c)$ mapping head $h$ and channel $c$ to flat index, $h^* \in [0, m_h]$ and $c^* \in [0, m_d]$ defining active dimensions. This construction preserves group-aware permutation structure: for heads $h, h' \in \mathcal{G}_g$ belonging to group $g$, $\mathbf{I}_m[\phi(h,\cdot)] = \mathbf{I}_m[\phi(h',\cdot)]$, and maintains head channel consistency: $\mathbf{I}_m[\phi(\cdot,c)]$ is uniform across all heads for each channel $c$.

\paragraph{Forward Pass.} The dynamic Mamba layer processes input through projection matrices following masked layer normalization. First, we apply the embedding mask to the layer norm output:
\begin{equation}
y_{\text{ln}} = \mathcal{M}_{\text{emb}}(\text{LN}(y))
\end{equation}
Then, projections are computed from the masked normalized input:
\begin{equation}
\begin{split}
z &= \mathbf{W}_z \cdot y_{\text{ln}}, \quad x = \mathbf{W}_x \cdot y_{\text{ln}}, \\
B &= \mathbf{W}_B \cdot y_{\text{ln}}, \quad C = \mathbf{W}_C \cdot y_{\text{ln}}, \quad d_t = \mathbf{W}_{dt} \cdot y_{\text{ln}}
\end{split}
\end{equation}
where $\mathbf{W}_z, \mathbf{W}_x \in \mathbb{R}^{(m_h \cdot m_d) \times d_e}$, $\mathbf{W}_B, \mathbf{W}_C \in \mathbb{R}^{(g \cdot d_s) \times d_e}$, and $\mathbf{W}_{dt} \in \mathbb{R}^{m_h \times d_e}$. Here, $d_e$ is the embedding dimension, $m_h$ denotes Mamba heads, $m_d$ is the head channel dimension, $g$ represents the number of Mamba groups, and $d_s$ is the SSM state dimension.

We apply the Mamba-specific mask to $z$, $x$, and $d_t$:
\begin{equation}
\begin{aligned}
z &\leftarrow \mathcal{M}_{\text{mamba}}(z), \\
x &\leftarrow \mathcal{M}_{\text{mamba}}(x), \\
d_t &\leftarrow \mathcal{M}_{\text{mamba}}(d_t)
\end{aligned}
\end{equation}

The intermediate activations $x$, $B$, and $C$ undergo causal convolution:
\begin{equation}
\hat{x} = \text{conv1d}(x), \quad \hat{B} = \text{conv1d}(B), \quad \hat{C} = \text{conv1d}(C)
\end{equation}
where the conv1d operation on $\hat{x}$ implicitly respects the Mamba mask structure.

The selective state-space model update computes:
\begin{equation}
\tilde{y} = \text{SSM}(\hat{x}, \hat{B}, \hat{C}, \mathbf{A}, \mathbf{D}, d_t)
\end{equation}

Followed by gated RMSNorm and output projection:
\begin{equation}
y_{\text{pre}} = \mathbf{W}_O \cdot \text{RMSNorm}(\tilde{y} \odot \text{silu}(z))
\end{equation}
where $\mathbf{W}_O \in \mathbb{R}^{d_e \times (m_h \cdot m_d)}$.

Finally, both dynamic masks are applied to the layer output:
\begin{equation}
y_{\text{out}} = \mathcal{M}_{\text{emb}}(\mathcal{M}_{\text{mamba}}(y_{\text{pre}}))
\end{equation}

\subsubsection{Dynamic Attention}

For multi-head attention layers in the hybrid architecture, we apply head-wise and embedding dimension masking to control capacity.

\paragraph{Dynamic Attention Head Mask Operator.} The operator $\mathcal{M}_{\text{attn\_head}}$ applies to matrices where one dimension derives from attention heads $n_h$ or head dimension $d_h$. For a matrix $\mathbf{W} \in \mathbb{R}^{f(n_h, d_h) \times d_e}$ where $f(n_h, d_h) = n_h \times d_h$, the masked operation is:
\begin{equation}
\mathcal{M}_{\text{attn\_head}}(\mathbf{W}) = \mathbf{W} \odot (\mathbf{I}_a \otimes \mathbf{1}_{d_e})
\end{equation}
where $\mathbf{I}_a \in \{0,1\}^{n_h \cdot d_h}$ satisfies:
\begin{equation}
\mathbf{I}_a[\psi(n,d)] = \begin{cases}
1 & \text{if } n \leq n^* \text{ and } d \leq d^* \\
0 & \text{otherwise}
\end{cases}
\end{equation}
with $\psi(n,d)$ mapping head $n$ and head dimension $d$ to flat index, $n^* \in [0, n_h]$ and $d^* \in [0, d_h]$ defining active dimensions.

\paragraph{Forward Pass.} The dynamic attention layer processes input through masked layer normalization:
\begin{equation}
y_{\text{ln}} = \mathcal{M}_{\text{emb}}(\text{LN}(y))
\end{equation}

Projections for query, key, and value are computed as:
\begin{equation}
\begin{split}
\mathbf{Q} &= \mathcal{M}_{\text{attn\_head}}(\mathbf{W}_Q) \cdot y_{\text{ln}}, \\
\mathbf{K} &= \mathcal{M}_{\text{attn\_head}}(\mathbf{W}_K) \cdot y_{\text{ln}}, \\
\mathbf{V} &= \mathcal{M}_{\text{attn\_head}}(\mathbf{W}_V) \cdot y_{\text{ln}}
\end{split}
\end{equation}
where $\mathbf{W}_Q, \mathbf{W}_K, \mathbf{W}_V \in \mathbb{R}^{(n_h \cdot d_h) \times d_e}$, with $n_h$ denoting attention heads and $d_h$ the head dimension.

The attention computation follows:
\begin{equation}
\text{Attn} = \text{softmax}\left(\frac{\mathbf{Q} \mathbf{K}^T}{\sqrt{d_h}}\right) \mathbf{V}
\end{equation}

Followed by output projection:
\begin{equation}
y_{\text{pre}} = \mathcal{M}_{\text{emb}}(\mathbf{W}_O) \cdot \text{Attn}
\end{equation}
where $\mathbf{W}_O \in \mathbb{R}^{d_e \times (n_h \cdot d_h)}$.

Finally, both dynamic masks are applied to the layer output:
\begin{equation}
y_{\text{out}} = \mathcal{M}_{\text{emb}}(\mathcal{M}_{\text{attn\_head}}(y_{\text{pre}}))
\end{equation}

\subsubsection{Dynamic FFN}

This formulation is applicable to both traditional dense FFN and the FFN in each expert in an MoE layer. For these layers, we apply masking to both embedding and intermediate dimensions.

\paragraph{Dynamic FFN Mask Operator.} The operator $\mathcal{M}_{\text{ffn}}$ applies to matrices where one dimension derives from the FFN intermediate dimension $f_\delta$, where $\delta$ is the layer number corresponding to the current FFN, $1\le\delta\le N_{\text{F}}$. For a matrix $\mathbf{W} \in \mathbb{R}^{f_\delta \times d_e}$, the masked operation is:
\begin{equation}
\mathcal{M}_{\text{ffn}}(\mathbf{W}) = \mathbf{W} \odot (\mathbf{I}_f \otimes \mathbf{1}_{d_e})
\end{equation}
where $\mathbf{I}_f \in \{0,1\}^{f_\delta}$ with $\mathbf{I}_f[0:j] = 1$ and $\mathbf{I}_f[j+1:f_\delta] = 0$ for some $j \in [0, f_\delta]$. For matrices $\mathbf{W} \in \mathbb{R}^{d_e \times f_\delta}$, the mask broadcasts similarly.

\paragraph{Forward Pass.} The dynamic FFN layer processes input through masked layer normalization:
\begin{equation}
y_{\text{ln}} = \mathcal{M}_{\text{emb}}(\text{LN}(y))
\end{equation}

The first linear transformation with dynamic masking:
\begin{equation}
h = \mathcal{M}_{\text{ffn}}(\mathbf{W}_1) \cdot y_{\text{ln}}
\end{equation}
where $\mathbf{W}_1 \in \mathbb{R}^{f_\delta \times d_e}$ and $f_\delta$ is the intermediate dimension.

Followed by activation and second linear transformation:
\begin{equation}
y_{\text{pre}} = \mathcal{M}_{\text{emb}}(\mathbf{W}_2) \cdot \sigma(h)
\end{equation}
where $\mathbf{W}_2 \in \mathbb{R}^{d_e \times f_\delta}$ and $\sigma(\cdot)$ denotes the activation function.

Finally, both dynamic masks are applied to the layer output:
\begin{equation}
y_{\text{out}} = \mathcal{M}_{\text{emb}}(\mathcal{M}_{\text{ffn}}(y_{\text{pre}}))
\end{equation}

\subsubsection{Dynamic MoE}

For MoE layers, elasticity is realized by masking experts at the router level rather than modifying expert FFNs directly.

\paragraph{Dynamic Expert Mask Operator.} The router produces logits over all experts via a learned linear projection:
\begin{equation}
\boldsymbol{\Theta} = \mathbf{W}_{\text{router}} \cdot x \in \mathbb{R}^{B \times E},
\end{equation}
where $B$ is the number of tokens and $\mathbf{W}_{\text{router}} \in \mathbb{R}^{E \times d_e}$ is the router weight matrix. Given a target expert count $e(\ell)$ at layer $\ell$, and a global per-layer importance ranking of experts (from REAP~\cite{lasby2025reap}), we construct a binary expert mask $\mathbf{m}_{\text{expert}}$ that retains only the top $e(\ell)$ experts:
\begin{equation}
\mathbf{m}_{\text{expert}}[j] =
\begin{cases}
1 & \text{if expert } j \text{ is among the top } e(\ell) \\
0 & \text{otherwise.}
\end{cases}
\end{equation}
This mask is applied along the expert dimension of the router logits before the routing operation:
\begin{equation}
\boldsymbol{\Theta}' = \boldsymbol{\Theta} + \log \mathbf{m}_{\text{expert}},
\end{equation}
where masked experts receive $-\infty$ logits, preventing them from being selected.

\paragraph{Forward Pass.} The dynamic MoE layer first computes router logits and applies the expert mask before routing:
\begin{equation}
\begin{aligned}
x_{\text{in}} &= \text{LN}(y), \\
\boldsymbol{\Theta} &= \mathbf{W}_{\text{router}} \cdot x_{\text{in}}, \\
\boldsymbol{\Theta}' &= \boldsymbol{\Theta} + \log \mathbf{m}_{\text{expert}},
\end{aligned}
\end{equation}

The masked logits $\boldsymbol{\Theta}'$ are passed to the router's routing function, which performs TopK selection with load balancing:
\begin{equation}
\mathbf{scores}, \mathbf{routing\_map} = \text{routing}(\boldsymbol{\Theta}'),
\end{equation}
The routing function selects the top-$k$ experts per token from the masked logits, applies load-balancing losses (e.g., auxiliary loss, sequence-level load balancing), enforces capacity constraints, and returns routing weights and a routing map indicating token-to-expert assignments. Since only the top $e(\ell)$ experts have finite logits, at most $e(\ell)$ experts can be selected per token, realizing the budget constraint.

\subsubsection{Depth Adaptation.} Layer-wise depth adaptation is achieved through selective layer retention controlled by $\boldsymbol{\gamma}$. The set of active layers is:
\begin{equation}
\mathcal{A} = \{j \mid \gamma_j = 1, j \in [0, N-1]\}
\end{equation}
where $|\mathcal{A}| = N_{\text{target}}$ specifies the target model depth. Skipped layers are bypassed via residual connections:
\begin{equation}
y_{j+1} = \begin{cases}
y_j + \mathcal{D} \circ \mathcal{L}_j(y_j) & \text{if } \gamma_j = 1 \\
y_j & \text{if } \gamma_j = 0
\end{cases}
\end{equation}
This maintains signal propagation while reducing computation. For hybrid architectures, selective layer retention enables leveraging the complementary strengths of Mamba and attention components at different model scales.

\subsubsection{Mask Generation}

\paragraph{Mask Generation from Router Output.} The router outputs $\mathbf{\pi}^{(\texttt{axis})}$ are processed through Gumbel-Softmax to produce relaxed discrete selections, where $\texttt{axis} \in \{d_e, m_h, m_d, n_h, e, f\}$. The selected configuration index is determined by $\hat{a}_{\texttt{axis}} = \arg\max_i \boldsymbol{P}^{(\texttt{axis})}_i$, where $\boldsymbol{P}^{(\texttt{axis})}$ is the Gumbel-Softmax probability distribution. In homogeneous mode, if dimension $\texttt{axis}$ selects configuration index $\hat{a}_{\texttt{axis}}$, the corresponding target count is $c_{\hat{a}_{\texttt{axis}}}$ (e.g., number of active embedding channels, depth, or head counts per layer). The binary mask is then constructed by selecting the top $c_{\hat{a}_{\texttt{axis}}}$ components according to the importance-based ranking $\sigma^{(\texttt{axis})}$:
\begin{equation}
\mathbf{I}^{(\texttt{axis})} = \mathbf{I}[\sigma^{(\texttt{axis})}(j) \leq c_{\hat{a}_{\texttt{axis}}}], \quad j = 1, \ldots, \text{size}^{(\texttt{axis})}
\end{equation}

In heterogeneous mode, the router output is reshaped into per-layer selections: $\mathbf{\pi}^{(\texttt{axis})}$ is partitioned into $N_{\text{X}}$ segments of size $|\mathcal{X}|$, where each segment determines the configuration for one layer. Per-layer masks are constructed similarly, allowing each layer to have distinct compression ratios. For depth selection, if the router outputs $L_{\text{target}} \in [1, N]$, the top $L_{\text{target}}$ layers from the importance depth ranking are activated via $\gamma_j = 1$ for the selected layers.

The generated masks are then applied to the dynamic model operators $\mathcal{M}_{\text{emb}}, \mathcal{M}_{\text{mamba}}, \mathcal{M}_{\text{attn\_head}}, \mathcal{M}_{\text{ffn}}$, and depth retention coefficients $\boldsymbol{\gamma}$ as defined in the Dynamic Model Formulation section, enabling the model to dynamically adjust capacity.

\paragraph{Mask Integration Strategies.} The Gumbel-Softmax probabilities provide differentiable signals for router optimization. We support two mask integration modes:

\emph{Mode 1: Hard Selection via Argmax Logits.} The discrete selection is obtained by $\hat{i}_{\texttt{axis}} = \arg\max_i P^{(\texttt{axis})}_i$, and a hard mask is applied using the corresponding logit:
\begin{equation}
\mathbf{I}^{(\texttt{axis})}_{\text{train}} = \mathbf{\pi}^{(\texttt{axis})}_{\hat{i}_{\texttt{axis}}} \cdot \mathbf{I}_{\hat{i}_{\texttt{axis}}}
\end{equation}
This directly applies the mask from the selected configuration, scaled by its logit magnitude to provide task-relevant gradient signals.

\emph{Mode 2: Soft Masking via Probabilistic Combination.} Alternatively, masks from all candidate configurations are combined proportionally to their probabilities:
\begin{equation}
\mathbf{I}^{(\texttt{axis})}_{\text{train}} = \sum_i P^{(\texttt{axis})}_i \cdot \mathbf{I}_i
\end{equation}
During training, this soft mask is applied to the dynamic operators, allowing gradients to flow through all configuration options. At inference time, the discrete mask corresponding to $\hat{i}_{\texttt{axis}}$ from the argmax mode is used and the logit, $\mathbf{\pi}^{(\texttt{axis})}_{\hat{i}_{\texttt{axis}}}$ is set to 1. We use soft masking throughout this paper.

\subsection{Elastic Model Deployment}
\label{app:deployment-details}
A key advantage of the elastic architecture is the ability to extract multiple model variants from a single trained checkpoint without requiring separate training or fine-tuning. This is achieved through a learned slicing mechanism that leverages the router module trained during the elastic training phase.

After training converges, the router has learned optimal budget-aware decisions for every layer and component (attention heads, Mamba, FFN, embeddings). At deployment time, to extract a model for any target budget \(B\) that was seen during training, we invoke the router with the budget specification. The router's learned decisions are used to determine which components should be pruned from the full model. These components are then permanently removed (sliced out) from the checkpoint, effectively extracting a nested sub-network that corresponds to the desired parameter count.

Formally, given a trained full model with parameter set \(\Theta_{\max}\) and a target budget \(B \in \mathcal{B}\) (where \(\mathcal{B}\) is the set of budgets used during training), the router \(\mathcal{R}\) produces a pruning specification that identifies the parameters to retain. The sliced model parameters are then:
\[
\Theta_B = \{\theta \in \Theta_{\max} : \theta \text{ is retained for budget } B\}
\]
This zero-shot slicing operation is computationally negligible and produces an inference-ready model immediately, with no retraining, fine-tuning, or additional distillation required. Crucially, any budget \(B \in \mathcal{B}\)---whether the largest, smallest, or any intermediate size explored during training---can be deployed directly from the single full-model checkpoint.

The same slicing pipeline applies to the FP8 and NVFP4 elastic checkpoints described in Section~\ref{sec:quantization}: because QAD preserves the nested mask hierarchy, the 23B and 12B variants can be sliced zero-shot from the 30B quantized checkpoint with no additional calibration.

\end{document}